\documentclass[journal]{IEEEtran}
\usepackage[dvipsnames]{xcolor}
\usepackage{amsmath,amssymb,amsfonts}
\usepackage{algorithmic}
\usepackage{graphicx}
\usepackage{textcomp}
\usepackage{multirow}
\usepackage{bbold}
\usepackage{hyperref}
\hypersetup{
    colorlinks=true,
    linkcolor=blue,
    filecolor=magenta,      
    urlcolor=cyan,
    pdftitle={Overleaf Example},
    pdfpagemode=FullScreen,
    }
\usepackage[ruled,vlined]{algorithm2e}

\newcommand{\etal}{\emph{et al.}~}
\newcommand{\ie}{\emph{i.e.}}
\newcommand{\eg}{\emph{e.g.}}

\definecolor{forestgreen}{rgb}{0.13, 0.55, 0.13}

\begin{document}

\title{CNN Attention Guidance for Improved Orthopedics Radiographic Fracture Classification}

\author{Zhibin Liao, Kewen Liao, Haifeng Shen, Marouska F. van Boxel, Jasper Prijs, Ruurd L. Jaarsma, Job N. Doornberg, Anton van den Hengel, \and Johan W. Verjans
\thanks{Z. Liao and K. Liao have contributed equally to this work.}
\thanks{Z. Liao and A. van den Hengel are with the Australian Institute for Machine Learning, University of Adelaide, Adelaide SA 5005, Australia. E-mail: \{zhibin.liao, anton.vandenhengel\}@adelaide.edu.au.}
\thanks{K. Liao and H. Shen are with the HilstLab, Peter Faber Business School, Australian Catholic University, North Sydney NSW 2060, Australia. E-mail: \{kewen.liao,  haifeng.shen\}@acu.edu.au.}
\thanks{M. F. van Boxel is with the Department of Orthopaedic Surgery, Amsterdam UMC, Amsterdam 1105 AZ, the Netherlands, and the Orthopaedic Department, Flinders Medical Centre and Flinders University, Bedford Park SA 5042, Australia. E-mail: m.f.vanboxel@amsterdamumc.nl}
\thanks{J. Prijs and J. N. Doornberg are with the Department of Surgery and the Department of Orthopaedic Surgery, Groningen University Medical Centre, Groningen 9713 GZ, the Netherlands, and the Orthopaedic Department, Flinders Medical Centre and Flinders University, Bedford Park SA 5042, Australia. E-mail: \{j.prijs, j.n.doornberg\}@umcg.nl}
\thanks{R. L. Jaarsma is with the Orthopaedic Department, Flinders Medical Centre and Flinders University, Bedford Park SA 5042, Australia. E-mail: ruurd.jaarsma@flinders.edu.au}
\thanks{J. W. Verjans is with the Australian Institute for Machine Learning, University of Adelaide, Adelaide SA 5005, Australia, and the South Australian Health and Medical Research Institute, Adelaide SA 5000, Australia. E-mail: johan.verjans@adelaide.edu.au}
}

\maketitle

\begin{abstract}

Convolutional neural networks (CNNs) have gained significant popularity in orthopedic imaging in recent years due to their ability to solve fracture classification problems. 
A common criticism of CNNs is their opaque learning and reasoning process, making it difficult to trust machine diagnosis and the subsequent adoption of such algorithms in clinical setting. This is especially true when the CNN is trained with limited amount of medical data, which is a common issue as curating sufficiently large amount of annotated medical imaging data is a long and costly process. 
While interest has been devoted to explaining CNN learnt knowledge by visualizing network attention, the utilization of the visualized attention to improve network learning has been rarely investigated.
This paper explores the effectiveness of regularizing CNN network with human-provided attention guidance on where in the image the network should look for answering clues.
On two orthopedics radiographic fracture classification datasets, through extensive experiments we demonstrate that explicit human-guided attention indeed can direct correct network attention and consequently significantly improve classification performance. 
The development code for the proposed attention guidance is publicly available on \href{https://github.com/zhibinliao89/fracture_attention_guidance}{GitHub}.
\end{abstract}


\section{Introduction}
\label{sec:introduction}

In orthopedics, fracture diagnosis mainly relies on X-ray because of its clear advantages in availability, cost, and speed compared to MRI and CT~\cite{olczak2017artificial}.
A fracture may be subtle or occult on plain radiographs; hence, fracture detection can be a challenging task, representing up to 80\% of the missed diagnoses in an emergency department~\cite{guly2001diagnostic}.
Some patients with a clinical suspicion of a fracture but negative radiograph require further MRI or CT examinations for confirmation.
For an orthopedic surgeon to accurately appreciate fractures on radiographs, years of training is required and the accuracy is subject to the observer's experience.
Convolutional Neural Networks (CNNs), on the other hand, have been successfully applied to fracture detection and grading tasks in radiographs,~\cite{olczak2017artificial, lindsey2018deep, langerhuizen2020deep, ozkaya2020evaluation, olczak2020ankle} showing remarkable performance on par with clinical experts.

While the performance of CNN-powered Artificial Intelligence (AI) methods is encouraging, human understanding towards what knowledge is captured by CNNs is still elementary.
As a false diagnosis can lead to undesirable, potentially disastrous consequences to a patient's well-being, an understanding of the reasoning process behind machine's decisions is of great importance to medical practitioners and the general public, who may put their trust in AI doctors.
On the other hand, the success of CNNs also greatly relies on the voluminous annotated data samples (\ie, ImageNet~\cite{deng2009imagenet}), whereas preparing medical imaging data with appropriate annotation is a time-intensive and costly process~\cite{willemink2020preparing}.
This correlates to the observations~\cite{tajbakhsh2019surrogate, soffer2019convolutional, willemink2020preparing} that medical image analysis tasks are often limited by the scarcity of data, which can cause a learning model to overfit to training data and hence result in poor generalization performance of the model.
Therefore, it is important to understand what CNN learns when limited medical imaging data are used -- even if the predicted outcome of CNN appears to be accurate.

To unravel the knowledge captured by a classification CNN model, a common method is visualizing the CNN ``attention'' heatmaps, such as using the class-wise Class Activation Maps~\cite{zhou2016learning} (CAM) and Gradient-weighted CAM (Grad-CAM)~\cite{selvaraju2017grad}.
In the context of medical machine learning, visualized heatmaps can illustrate the lesion or tissue patterns that trigger the formation of an AI diagnosis, enabling human sense-making of AI's decisions in a visual grounding~\cite{deng2018visual} manner.
Such use cases can be found in a number of applications, \eg, detecting Alzheimer's disease~\cite{tang2019interpretable} from neuropathology images, classifying lung diseases~\cite{irvin2019chexpert} from Chest X-rays, and grading brain tumor from MRI~\cite{pereira2018automatic}.
Nevertheless, the grounded patterns may not always make sense.
In this work, we observe that a trained CNN classifier may make a prediction based on random salient patterns when the number of training samples is limited and even though the prediction outcome is correct (see Fig.~\ref{fig:scaphoid_example}-(b) and~\ref{fig:scaphoid_occult_example}-(b)).

If the visualized CNN attention is incorrect or inaccurate, humans could guide the network to correct that, but the challenge is how humans can provide such guidance.
Attention guidance is widely studied in high-level tasks such as image captioning and visual question answering (VQA) ~\cite{liu2016attention,das2017human,qiao2017exploring,li2018tell,zhang2019interpretable,selvaraju2019taking}, where annotated or mined localization information (\ie, object bounding boxes or segmentation masks) is used as the ground truth. 
However, attention guidance is rarely visited in image classification tasks. Most of these tasks use commonly benchmarked image classification datasets whose volumes are sufficient (some even excessive) to generate accurate classification network attention. On the other hand, in the case of orthopedics radiographic fracture classification, image samples are often limited or expensive to obtain and their annotations can be error-prone even for clinical experts such as judging subtle and occult fractures, let alone annotating their exact locations. Hence, human-provided attention guidance with a variety of annotation forms could become critical and effective as the clinical experts can use their knowledge accumulated over years of practice to guide the CNN model instead of letting the CNN to find label-correlated image patterns from scratch.

In this work, we explore the feasibility of guiding CNN attention for radiographic fracture classification tasks using clinical experts provided attention ground truth. Our proposed attention guidance technique is formed as a network regularization mechanism comprising two parts: 1) an elaborate modification of the widely used state-of-the-art CNN architecture (\eg, using ResNet~\cite{he2016deep} as the backbone model) to accept human guided attention, and 2) a novel attention regularization loss function to incorporate the accepted guidance signals during network training.
Extensive experiments on our proprietary scaphoid~\cite{langerhuizen2020deep} and ankle fracture~\cite{prijs2021development} classification datasets demonstrate that the proposed attention guidance method can lead to more accurately grounded fracture localization and better classification performance at the same time. The main contributions of this work are summarized as follows.
\begin{itemize}
    \item To tackle CNN training with limited medical image samples, we propose a novel attention guidance method that incorporates human-guided regularization to reinforce the dependency between CNN class-wise attention and class prediction and consequently achieve significantly improved results. %
    \item To address the difficulty in fracture annotation, we leverage and validate focus region annotations -- annotating an area to be focused on by the network using styles of scribble, bounding box or segmentation mask instead of the need to reveal an exact fracture location.  %
    \item The proposed method demonstrates consistent and superior prediction performance over baselines (ranging from 5\% to 10\% depending on experiment settings) through extensive experiments across two challenging orthopedics fracture classification datasets.
    \item The proposed method provides end-to-end training and inference pipelines which enable large-size medical images to be accurately classified according to local features without the need to explicitly manually crop or detect the responsible local patches first.
\end{itemize}

\section{Literature Review}

\subsection{Orthopedics Fracture Classification Using CNNs}

Recent studies in orthopedics~\cite{olczak2017artificial, lindsey2018deep, langerhuizen2020deep, ozkaya2020evaluation, olczak2020ankle} have demonstrated that deep CNNs can be used to solve fracture classification problem with expert level performance.
Since the radiographic images are often in a high spatial resolution, they are often resized to a much smaller spatial size (\eg, $256 \times 256$)~\cite{olczak2017artificial, olczak2020ankle} to be compatible with a pretrained classification CNN, but this convenience is at the risk of losing detailed local patterns (see the illustration in Fig.~1 of \cite{olczak2017artificial}).
Taking scaphoid fracture as an example, the skewed ratio of the scaphoid bone versus the full wrist radiograph makes it difficult to appreciate the fracture after resizing, hence manual cropping of scaphoid is necessary~\cite{langerhuizen2020deep, ozkaya2020evaluation}.
However, the cropping operation harms the automation at the inference stage, reducing the practicability in clinical use.
Without heavily compromising the spatial resolution or applying manual cropping, Lindsay~\etal~\cite{lindsey2018deep} proposed using a U-Net~\cite{ronneberger2015u} segmentation network with an additional classification module to allow accurate and simultaneous fracture localization and classification on a large spatial size (\ie, $1024 \times 512$) input. 
Although the localization property increases the trustworthiness of the model, the model's classification prediction has no dependency on such localization. In comparison to~\cite{lindsey2018deep}, our proposal intentionally makes the model prediction fully dependent on the localized CNN attention.
\subsection{CNN Visualization}
One way to visualize a CNN's decision-making process is to treat it as a black box and apply model-agnostic manipulation to the input images~\cite{ribeiro2016should, fong2017interpretable} and then monitor the change in the prediction response to associate image patterns with classification decisions.
An arguably more popular way is by visualizing model attention.
For visualizing and explaining classification CNNs, Zhou~\etal~\cite{zhou2016learning} proposed Class Activation Mapping (CAM) which can provide object localization maps as a by-product of the feed-forward inference pass, with the catch that the network contains a Global Average Pooling (GAP)~\cite{lin2013network} layer followed by a softmax classification layer.
Such requirement is mostly satisfied by the state-of-the-art CNNs such as the ResNets~\cite{he2016deep} and GoogLeNets~\cite{szegedy2015going}.
More recent variations of CAM, such as Gradient-weighted CAM (Grad-CAM)~\cite{selvaraju2017grad} and Guided Backprop~\cite{springenberg2014striving} utilize the gradient computation to allow the generalization of the localization map beyond classification networks and with fine-grained details.

\subsection{CNN Attention Mechanism}
\label{sec: attMech}
Attention mechanism is frequently explored in solving vision and language (V\&L) tasks such as image captioning and VQA~\cite{xu2015show, xu2016ask, yang2016stacked, lu2016hierarchical}. 
A typical attention mechanism adds trainable softmax activated modules to a V\&L solver network to control the interaction between the language and vision modalities, achieved by weighting the importance among feature elements.

In the medical imaging analysis domain, attention mechanism has been widely explored in multi-imaging-modality or view feature fusion~\cite{zhang2021multi, yang2020simultaneous}, spatial feature selection~\cite{yuan2020sara, schlemper2019attention}, and designing attention-aware CNN architecture~\cite{liu2019automatic} with applications from image diagnosis and segmentation to image quality enhancement.
It should be noted that unlike the human-provided CNN attention guidance proposed in this paper, the attention mechanism aims at learning the attention without external supervision. Hence, it solely relies on task-related training signals to form the attention, where the semantic correctness of the generated attention localization and the effectiveness in model performance improvement are limited by data sufficiency. In fact, we have also observed this in our experiments of comparing the effectiveness of our proposed attention guidance method with several state-of-the-art attention mechanisms (see Table~\ref{tab:scaphoid_table} and Table~\ref{tab:ankle_table}).

\subsection{CNN Attention Guidance}
Different from the CNN attention mechanism, the CNN attention guidance is mainly provided by a human.
Similar to solving an object detection or a semantic segmentation task, CNN attention guidance is achievable by promoting spatial activation at the targeted regions, where ground truth can be scribbles~\cite{shen2021human}, object bounding boxes~\cite{zhang2019interpretable}, segmentation masks~\cite{li2018tell}, or annotated human attention~\cite{judd2009learning, das2017human, selvaraju2019taking}.
In comparison to bounding box and segmentation annotations, collecting authentic human attention is laborious and expensive which often requires specialized equipment such as an eye-tracking device. 
In addition, it has been shown that the attention generated from the VQA attention models does not necessarily align with human attention~\cite{das2017human}. On the other hand, using human provided attention guidance has been shown to yield a positive performance gain for image captioning~\cite{liu2016attention} and VQA~\cite{qiao2017exploring} tasks. %
For solving a classification problem, the attention guidance is rarely explored as the localization property of CAM derived from the ImageNet trained backbones is remarkably accurate~\cite{zhou2016learning}.
Nevertheless, in a medical image classification setting, the ImageNet trained CAM is often inaccurate and often there is a lack of sufficient amount of labeled samples for properly rectifying it; hence in this work, we attempt to guide the attention formation of the CNNs by a human for medical imaging tasks.

\begin{table}[!htbp]
    \centering
    \caption{The scaphoid fracture dataset composition. In each table cell, the numbers are noted ``\# studies (corresponding \# images)''.
    }
    \resizebox{0.8\columnwidth}{!}{
        \begin{tabular}{|c|c|c|c|c|c|}
        \hline
        & Fracture & No Fracture & Total \\
        \hline
        Training Set & 80 (241) & 80 (238) & 160 (479) \\
        \hline
        Validation Set & 20 (59) & 20 (60) & 40 (119) \\
        \hline
        Test Set & 50 (150)  & 50 (150) & 100 (300) \\
        \hline
        \end{tabular}
    }
    \label{tab:scaphoid_dataset}
\end{table}

\section{Datasets and Annotations}
\label{sec:datasets}

\subsection{Scaphoid Fracture Dataset}
The first investigated dataset is a challenging scaphoid fracture classification dataset~\cite{langerhuizen2020deep} collected at the Flinders Medical Centre (FMC), Adelaide, Australia. 
The scaphoid is one of the eight carpal bones of the wrist. Fractures of the scaphoid are the most common carpal bone injuries~\cite{beasley2003beasley}. Diagnosis can be challenging because one out of six scaphoid fractures is missed on initial radiographs. Hence patients with a clinical suspicion and negative radiographs require further imaging~\cite{mallee2011comparison}.
The explored scaphoid fracture dataset~\cite{langerhuizen2020deep} contains 300 radiographic studies, 150 having scaphoid fractures and the rest having no fracture.
Each study has a number of 2 to 4 anonymized images including the posteroanterior (PA), PA ulnar deviation and uptilt, oblique, and lateral views, resulting in a total of 1197 images.
We followed the same training and test split as was used by Langerhuizen~\etal~\cite{langerhuizen2020deep}, where 200 studies were assigned to the training set and 100 studies to the test set.
We further selected 40 studies randomly from the training set as a validation set while maintaining a 50:50 ratio of the fracture and no fracture population in each set.

Each image of a study inherits the fracture label assigned to the study, however there is no guarantee that the fracture can be solely appreciated correctly from any individual view, \ie, a fracture may be occult meaning it is not visible on any view or, in some cases, only visible on one view.
As a result, the diagnosis was confirmed with a corresponding CT or MRI scan.
From a statistical standpoint, out of the 150 fracture studies, 127 cases can be confirmed on the radiographs by experts and the rest 23 fracture cases were only visible on MRI or CT. 
To facilitate this work, we annotated auxiliary information for the scaphoid fracture dataset.
First, we made additional view labels which are used to remove the lateral view images.
This is because fractures are difficult to appreciate in the lateral view as the scaphoid is partially overlapped with the capitate, pisiform, triquetrum and lunate bones.
We eventually reached 479 images for training, 119 for validation, and 300 images for testing (see data statistics in Table~\ref{tab:scaphoid_dataset}).
Second, we made additional focus region annotations in scribble, bounding box, and segmentation styles.
Given our aim to detect scaphoid fractures, the respective annotation is determined as a vertical scaphoid mid-line scribble, a tight bounding box around the scaphoid bone, and a scaphoid bone contour tracing (see ground truths (GTs) in Fig.~\ref{fig:scaphoid_example}-(c) to (e)).

\subsection{Ankle Fracture Dataset}

\begin{table}[!htbp]
    \centering
    \caption{The ankle fracture dataset data composition. In each table cell, the numbers are noted ``\# studies (corresponding \# images)''.}
    \resizebox{\columnwidth}{!}{
        \begin{tabular}{|c|c|c|c|c|c|}
        \hline
        & Weber A & Weber B & Weber C & No Frac. & Total \\
        \hline
        Training Set & 33 (50) & 136 (187) & 16 (39) & 253 (406) & 438 (682) \\
        \hline 
        Validation Set & 10 (16) & 10 (14) & 10 (20) & 10 (17) & 40 (67) \\
        \hline
        Test Set &  25 (75) & 25 (75) & 25 (75) & 25 (75) & 100 (300) \\
        \hline
        External Test Set &  11 (23) & 62 (140) & 17 (34) & 67 (143) & 157 (340) \\
        \hline
        \end{tabular}
    }
    \label{tab:ankle_dataset}
\end{table}

The second investigated dataset~\cite{prijs2021development} is an ankle fracture dataset also collected at the FMC.
The dataset contains a total of 1,049 ankle radiographic images from 578 patient studies.
The fractures of interest in this dataset are the lateral malleolar fractures corresponding to the level of the fracture in relation to the ankle joint, measured as type A, B, or C by the Danis-Weber fracture classification system~\cite{hunter2000radiologic} (abbreviated as Weber for the rest of the paper), see the example in Fig.~\ref{fig:fracture_example}-(a).
We also collected 340 ankle radiographic images of 157 patient studies from the University Medical Centre Groningen, Groningen, the Netherlands for external validation.
The ankle dataset composition is depicted in Table~\ref{tab:ankle_dataset}.
As an overview, we found the original data exhibits a natural bias towards the Weber B fracture and healthy no fracture groups.
For the images in the training and validation sets, each study may include 1 to 3 training images, \ie, any of the anteroposterior (AP), mortise, and lateral views for the ankle joint.
For testing purpose, we compose a stratified test set with a composition of an equal 25-study volume for each class, resulting a 100-study test set each with all three views available.
In addition to the annotation of fracture classification labels, the dataset has been annotated with extra fibula segmentation masks.
From these masks, we computed the bounding boxes and scribbles (mimicking the vertical mid-line scribble using the skeletonization function provided by scikit-image image processing library~\cite{scikit-image}) as the additional explored focus region annotation styles.

\section{Methodology}

\begin{figure*}[!htbp]
    \centering
    \includegraphics[width=0.9\textwidth]{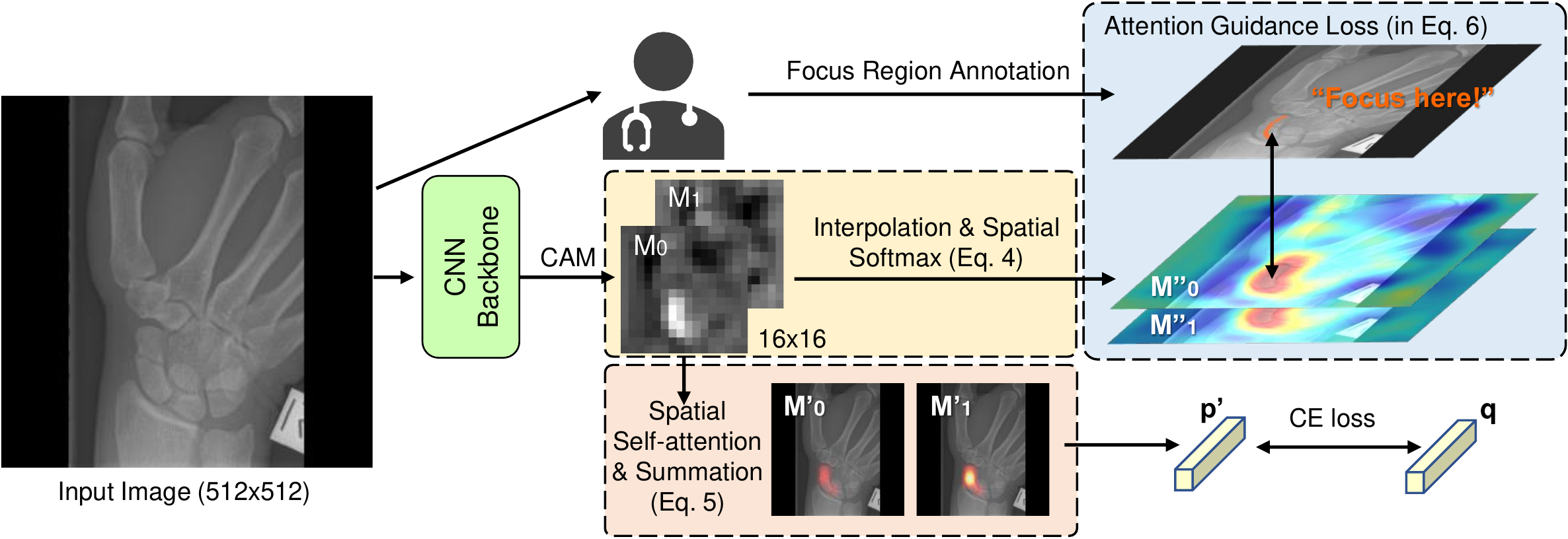}
    \caption{The schematic of the proposed CNN attention guidance method. We introduce three components (in dashed boxes) to incorporate human provided attention guidance in CNN classifier training. $\mathbf{M}_c$'s ($c \in \{0, 1\}$) denote the CAM attention maps and are transformed into $\mathbf{M''}_c$'s to be compared with the human provided focus region annotation for computing the proposed attention guidance loss. $\mathbf{M}_c$'s are also transformed to $\mathbf{M'}_c$'s which can be summed to a class probability prediction $\mathbf{p'}$ to be compared with the ground truth label $\mathbf{q}$ for computing the classification loss.
    }

    \label{fig:overview}
\end{figure*}

The proposed attention guidance method (see Fig.~\ref{fig:overview}) is achieved by regularizing Class Activation Maps (CAM) activation to follow the focus region guidance targets.
The intuition behind is that the activation values on CAM are directly summed to compute the logits for the final classification prediction; hence a pixel with large activation value for a CAM class not only indicates that a pattern is detected at the corresponding spatial region but also pushes up the confidence of the class presence.
Even though we cannot describe the pattern (\ie, how does a fracture look like) to CNN, promoting CAM to follow the focus region can guide the CNN to discover the pattern at the pointed region and associate that with the designated classification class at the same time.

In our formulation, we choose CAM over Grad-CAM as CAM can be computed during the same feed-forward phase and the final network prediction depends on the bias shifted CAM attention (\ie, see Eq.~(\ref{eq:cam})).

\subsection{Class Activation Maps}
\label{sec:cam}
The visualization of CAM~\cite{zhou2016learning} relies on the Global Average Pooling (GAP) layer, which is widely used in the state-of-the-art CNNs such as the GoogLeNets~\cite{szegedy2015going} and ResNets~\cite{he2016deep}.
In a nutshell, GAP collapses the spatial dimensions of the last convolution output maps, generating a channel-width long image encoding for the final softmax classification layer. 
Assuming that the last convolutional feature map tensor $\mathbf{F}$ has the size of $H \times W \times K$ and $\mathbf{F}_{ijk}$ denotes an element (pixel) in $\mathbf{F}$ with respect to the subscripted indices, the GAP operation can be written as:
\begin{equation}
    \mathbf{f} = \frac{1}{H \times W} \sum_{i \in H} \sum_{j \in W} \mathbf{F}_{ij},
\end{equation}
where $\mathbf{f} \in \mathbb{R}^{K}$ is the generated image encoding vector.
Let us also denote the number of classes as $C$, the weight matrix of the classification layer as $\mathbf{W} \in \mathbb{R}^{K \times C}$ and bias as $\mathbf{b} \in \mathbb{R}^{C}$, the computation of logits $\mathbf{z}$ and class probability distribution $\mathbf{p} \in \mathbb{R}^C$ are carried out as:
\begin{align}
    \mathbf{z} & = \mathbf{W}^{\intercal} \mathbf{f} + \mathbf{b},  \text{ and, } \mathbf{p} = \text{softmax}(\mathbf{z}).
\end{align}
Using $\mathbf{M} \in H \times W \times C$ to denote the tensor of CAM computed attention maps, the computation of a single map $\mathbf{M}_c$ is by removing GAP and then directly weighting $\mathbf{F}_{k} \in \mathbb{R}^{H \times W}$ by the corresponding weight elements in $\mathbf{W}_c$:
\begin{equation}
    \mathbf{M}_c = \sum_{k \in K} \mathbf{W}_{kc} \cdot \mathbf{F}_{k},
\label{eq:cam}
\end{equation}
and $\mathbf{z}$ can be rewritten as $\mathbf{z} = \frac{1}{H \times W} \sum_{i \in H} \sum_{i \in W} \mathbf{M} + \mathbf{b}$, showing that the classification prediction $\mathbf{p}$ is solely dependent on the bias shifted CAM.
Finally, for visualizing $\mathbf{M}_c$ as CAM~\cite{zhou2016learning}, it is common to: 1) up-sample $\mathbf{M}_c$ to the input size using the bicubic interpolation function $g$; 2) normalize the values by min($g(\mathbf{M}_c)$) and max($g(\mathbf{M}_c)$); and 3) overlay $g(\mathbf{M}_c)$ to the input image as a heatmap.

\subsection{Attention Guidance By CAM Regularization}

Let a focus region ground truth mask $\mathbf{S}_c$ be filled with 1s inside the annotation boundary and the rest unannotated pixels as 0s.
To use $\mathbf{S}_c$ as a supervision target for CAM, we again up-sample $\mathbf{M}_c$ to the input image size via function $g$ and then apply the softmax function over the spatial dimensions for each class, which achieves a differentiable soft-attention~\cite{xu2015show} that is compatible with the binary targets $\mathbf{S}_c$:
\begin{equation}
    \mathbf{M''}_c = \text{softmax}(g(\mathbf{M}_c + \mathbf{b}_c)),
\label{eq:up_sampled_cam}    
\end{equation}
where $\mathbf{b}_c$ is added to maintain the original decision dependency.
The soft-attention creates relativity in the spatial dimensions. 

Next unlike attention mechanisms, in our formulation there is no dedicated computation module to generate the attention values. Instead, the CAM activation values themselves can be used for the computation, which can be thought as a barebone self-attention~\cite{vaswani2017attention} in spatial dimensions without any additional model parameters. We incorporate the spatial self-attention procedure on the original scale to compute the CAM replacement attention maps $\mathbf{M}'_c$ and the new classification prediction $\mathbf{p'}$: 
\begin{align}
    \mathbf{M}'_c = \text{softmax}(\mathbf{M}_c + \mathbf{b}_c) \cdot (\mathbf{M}_c + \mathbf{b}_c), \nonumber\\
    \mathbf{z}' =  \sum_{i \in H} \sum_{j \in W} \mathbf{M}' \text{, and, } \mathbf{p}' = \text{softmax}(\mathbf{z}'),
\end{align}
where spatial self-attention replaces the functionality of GAP, assigning an unequal membership to every pixels on a single CAM activation map. %

Finally, the supervising loss for the network is a composition of two terms: the classification cross-entropy (CE) loss and the regularizing attention guidance loss in the form of a binary cross entropy (BCE): 
\begin{equation}
    \ell = \mathcal{H}(\mathbf{p'}, \mathbf{q}) + \frac{\lambda}{N} \sum_{i, j, c} \mathbb{1}(\mathbf{S}_{ijc} \neq 0)\mathcal{H}(\mathbf{M''}_{ijc}, \mathbf{S}_{ijc}),
\label{eq:loss}
\end{equation}
where $\mathcal{H}(., .)$ denotes the CE function, $\lambda$ denotes a weighting parameter to control regularization influence, $\mathbf{q}$ denotes the classification one-hot label vector, $\mathbb{1}(.)$ denotes the indicator function and $N = \sum_{i, j, c} \mathbb{1}(\mathbf{S}_{ijc} \neq 0)$ normalizes the term with respect to the annotated pixels; the BCE in Eq.~(\ref{eq:loss}) does not force all pixels outside the focus region to go zero, as this is implicitly carried out by the softmax normalization.

\subsection{Unified Focus Region}

It can be seen from Eq. (6) that the loss function facilitates an individual focus region ground truth to be supplied for each class $c$.
Nevertheless, we found it difficult for a human annotator to give a semantically meaningful and accurate class-wise focus region for every possible class. 
For instance, when encountering the healthy no fracture or subtle/occult fracture cases, only the target bone is annotatable.
To alleviate the annotation difficulty, we apply a unified focus region to all classes, i.e., $\forall c \neq c': \mathbf{S}_{ijc} = \mathbf{S}_{ijc'}$, where $c'$ denotes a different class other than $c$.

\begin{figure*}[!htbp]
    \centering
    \resizebox{0.9\textwidth}{!}{%
    \begin{tabular}{cc}
    \includegraphics[width=0.5\textwidth]{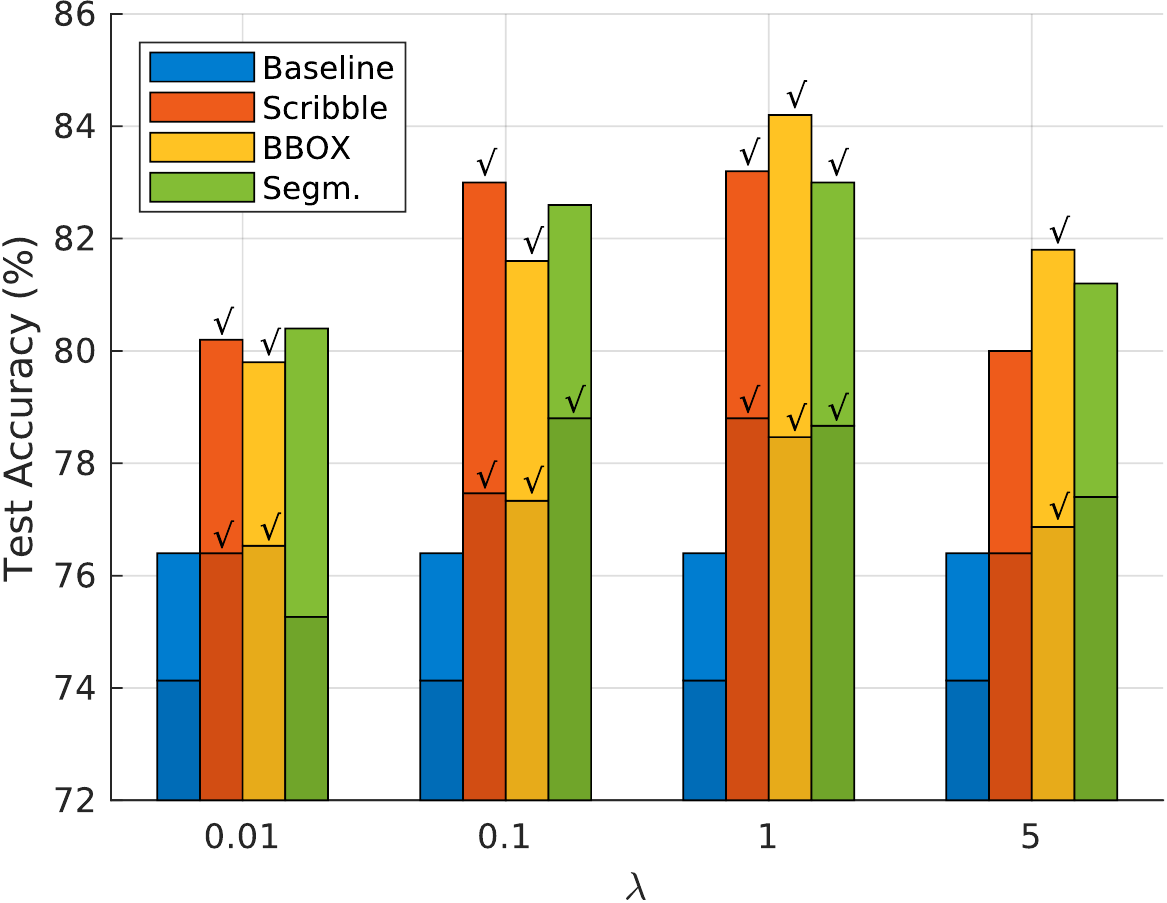}
    & 
    \includegraphics[width=0.5\textwidth]{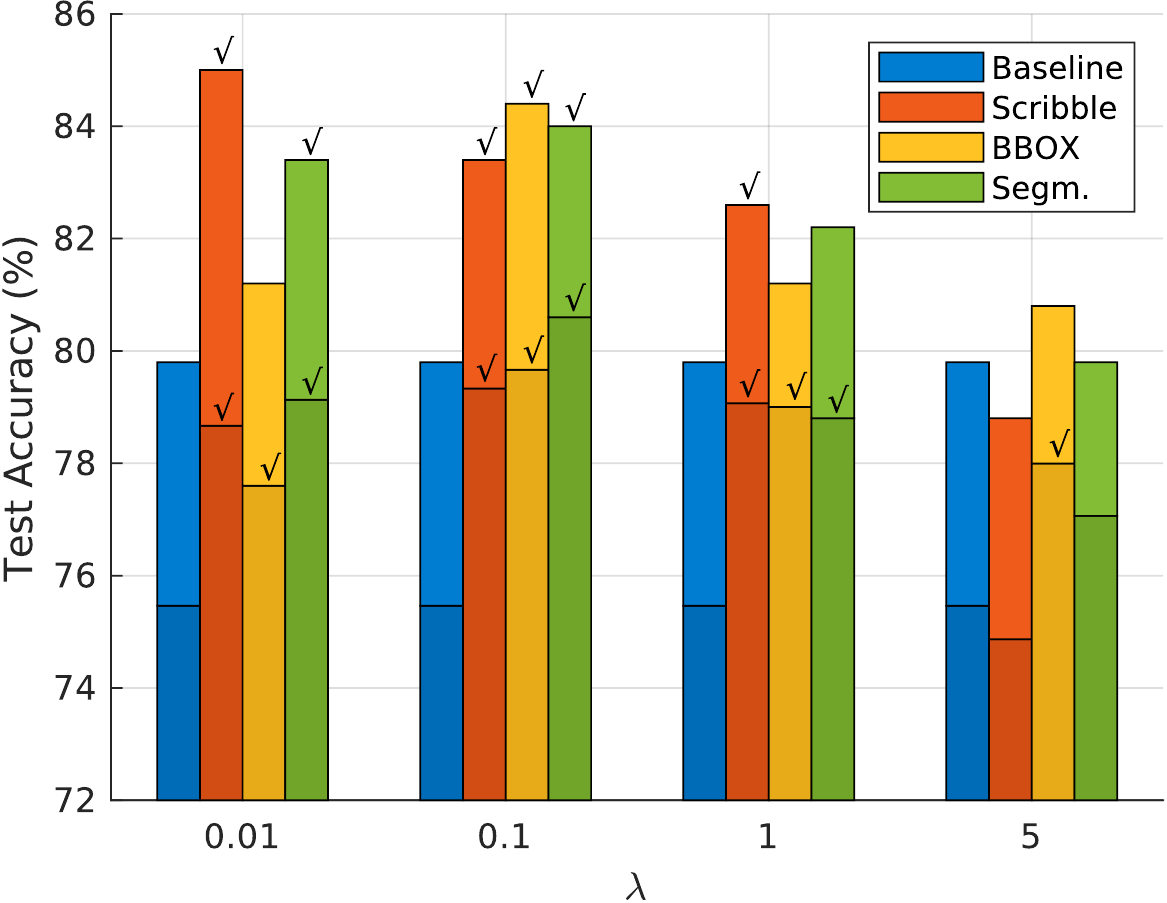} \\
    (a) Input size $512 \times 512$ & (b) Input size $768 \times 768$ \\
    \end{tabular}
    }
    \caption{The scaphoid fracture result comparison as a function of the $\lambda$ assigned to the proposed attention regularization loss in Eq.~(\ref{eq:loss}), supervised using either scribble, bounding box (BBOX), or segmentation (Segm.) focus region styles.
    For each bar, the darker bottom segment indicates the image-wise performance and the lighter top segment indicates the study-wise performance. Ticks denote statistically significant results over the baseline.}
    \label{fig:scaphoid_result}
\end{figure*}

\begin{table*}[!htbp]
    \caption{The scaphoid fracture dataset experiment results in classification accuracy (\%). Each table cell shows the statistical result computed by five repetitive runs of the same setting. The highest classification performance is highlighted for each column.}
    \centering
    \resizebox{\textwidth}{!}{
    \begin{tabular}{|c|c|c|c|c|c|c|c|c|c|c|c|c|}
        \hline
        \multirow{2}{*}{Methods}	&	\multicolumn{6}{c|}{s-512}											&	\multicolumn{6}{c|}{s-768}											\\ \cline{2-13}
	&	\multicolumn{3}{c|}{Image-wise}			&			\multicolumn{3}{c|}{Study-wise}					&	\multicolumn{3}{c|}{Image-wise}					&	\multicolumn{3}{c|}{Study-wise}					\\ \hline
Baseline	& \multicolumn{3}{c|}{$	74.1 \pm 1.3					$} & \multicolumn{3}{c|}{$	76.4 \pm 0.9					$} & \multicolumn{3}{c|}{$	75.5 \pm 1.2					$} & \multicolumn{3}{c|}{$	79.8 \pm 2.1					$} \\ \hline
L1 Reg. ($\lambda$=1e-8)	& \multicolumn{3}{c|}{$	76.4 \pm 2.0					$} & \multicolumn{3}{c|}{$	80.0 \pm 2.4					$} & \multicolumn{3}{c|}{$	76.1 \pm 4.0					$} & \multicolumn{3}{c|}{$	80.0 \pm 4.0					$} \\ \hline
L2 Reg. ($\lambda$=1e-8)	& \multicolumn{3}{c|}{$	76.7 \pm 2.1					$} & \multicolumn{3}{c|}{$	78.4 \pm 3.4					$} & \multicolumn{3}{c|}{$	78.1 \pm 2.0					$} & \multicolumn{3}{c|}{$	82.0 \pm 2.5					$} \\ \hline
Residual Attn.~\cite{wang2017residual}	& \multicolumn{3}{c|}{$	75.3 \pm 3.4					$} & \multicolumn{3}{c|}{$	78.8 \pm 2.4					$} & \multicolumn{3}{c|}{$	75.6 \pm 1.7					$} & \multicolumn{3}{c|}{$	78.6 \pm 2.9					$} \\ \hline
Self Attn.~\cite{zhang2019self}	& \multicolumn{3}{c|}{$	75.5 \pm 1.7					$} & \multicolumn{3}{c|}{$	79.2 \pm 1.8					$} & \multicolumn{3}{c|}{$	78.1 \pm 1.6					$} & \multicolumn{3}{c|}{$	81.6 \pm 2.5					$} \\ \hline
Pyramid Attn.~\cite{li2018pyramid}	& \multicolumn{3}{c|}{$	75.7 \pm 0.6					$} & \multicolumn{3}{c|}{$	79.2 \pm 2.7					$} & \multicolumn{3}{c|}{$	73.7 \pm 1.5					$} & \multicolumn{3}{c|}{$	78.2 \pm 3.6					$} \\ \hline
Attn. Guidance	&	Scribble	&	BBOX	&	Segm.	&	Scribble	&	BBOX	&	Segm.	&	Scribble	&	BBOX	&	Segm.	&	Scribble	&	BBOX	&	Segm.	\\ \hline
$\lambda$ = 0	& \multicolumn{3}{c|}{$			76.1 \pm 0.8			$} & \multicolumn{3}{c|}{$			78.8 \pm 2.2			$} & \multicolumn{3}{c|}{$			78.1 \pm 4.2			$} & \multicolumn{3}{c|}{$			81.6 \pm 3.8			$} \\ \hline
$\lambda$ = 0.01	& $	76.4 \pm 1.7	$ & $	76.5 \pm 1.9	$ & $	75.3 \pm 1.9	$ & $	80.2 \pm 0.8	$ & $	79.8 \pm 1.8	$ & $	80.4 \pm 2.2	$ & $	78.7 \pm 1.2	$ & $	77.6 \pm 1.5	$ & $	79.1 \pm 0.5	$ & $	\bf 85.0 \pm 1.4	$ & $	81.2 \pm 2.4	$ & $	83.4 \pm 1.5	$ \\ \hline
$\lambda$ = 0.1	& $	77.5 \pm 1.7	$ & $	77.3 \pm 1.4	$ & $	78.8 \pm 1.2	$ & $	83.0 \pm 2.0	$ & $	81.6 \pm 1.8	$ & $	82.6 \pm 3.0	$ & $	\bf 79.3 \pm 1.2	$ & $	\bf 79.7 \pm 1.6	$ & $	\bf 80.6 \pm 1.2	$ & $	83.4 \pm 2.0	$ & $	\bf 84.4 \pm 2.1	$ & $	\bf 84.0 \pm 1.6	$ \\ \hline
$\lambda$ = 1	& $	\bf 78.8 \pm 1.6	$ & $	\bf 78.5 \pm 1.4	$ & $	\bf 78.7 \pm 0.9	$ & $	\bf 83.2 \pm 2.2	$ & $	\bf 84.2 \pm 0.8	$ & $	\bf 83.0 \pm 2.0	$ & $	79.1 \pm 0.9	$ & $	79.0 \pm 1.3	$ & $	78.8 \pm 1.7	$ & $	82.6 \pm 1.5	$ & $	81.2 \pm 1.1	$ & $	82.2 \pm 1.5	$ \\ \hline
$\lambda$ = 5	& $	76.4 \pm 2.3	$ & $	76.9 \pm 1.3	$ & $	77.4 \pm 2.0	$ & $	80.0 \pm 4.9	$ & $	81.8 \pm 1.3	$ & $	81.2 \pm 2.4	$ & $	74.9 \pm 2.3	$ & $	78.0 \pm 1.2	$ & $	77.1 \pm 1.4	$ & $	78.8 \pm 2.2	$ & $	80.8 \pm 3.0	$ & $	79.8 \pm 2.4	$ 

        \\ \hline
    \end{tabular}
    }
    \label{tab:scaphoid_table}
\end{table*}

\section{Experiments}

In this section, we report experimental results for the two fracture datasets described in~Sec.~\ref{sec:datasets}.
The shared experiment settings are: 1) all CNNs are trained by using stochastic gradient descend (SGD) with 0.9 momentum and 5e-4 weight decay; 2) training lasts 100 epochs with linear learning rate decay; 3) the reported results are computed on the respective test set, trained using the combined training and validation sets and the training parameters were initially tuned on the validation set; 4) each obtained result is repetitively run for five instances (with random initialization) to compute the statistical performance; and 5) all baseline models mentioned in this section are the respective CNNs without using the proposed attention guidance.

\subsection{Scaphoid Fracture Classification}
\label{sec:scaphoid_exp}

{\bf Experiment setting:}
for the scaphoid fracture dataset, we mainly explore the effectiveness of attention guidance between the scribble, bounding box, and segmentation styles, under $512 \times 512$ or $768 \times 768$ pixel sizes (short noted as s-512 or s-768 for clarity).
We show both the image-wise and study-wise classification accuracy performance.
The study-wise accuracy is computed by averaging the view-wise probability predictions of a study and then selecting the class with the maximum probability.

{\bf Network specification:} 
we use the pretrained ResNet-50~\cite{he2016deep} model provided by PyTorch as the CNN backbone, removing the GAP and softmax layers. 
The ResNet-50 backbone contains 50 layers in the order of one convolution layer, one max-pooling layer, and the rest 48 convolution layers encapsulated in four residual blocks.
The output of the backbone is an $\mathbf{F} \in \mathbb{R}^{16 \times 16 \times 2048}$ for s-512 or an $\mathbf{F} \in \mathbb{R}^{24 \times 24 \times 2048}$ for s-768.
The batch size is set to 16 for s-512 and 8 for s-768 to be trainable on a single 11 GB Nvidia RTX 2080 Ti GPU.
The linear learning rate starts with 1e-3 and ends at 1e-5.
The radiographs are first padded to square size. 
The transformation for the input images is a sequence of random cropping, horizontal flipping, image rotation (\ie, up to $\pm 30$ degrees), before they are finally resized to the network input size.

{\bf Result Interpretation:}
the corresponding numerical results are shown in Table~\ref{tab:scaphoid_table} and the attention guidance experiment results as a function of $\lambda$ and by focus region types are also visualized in Fig.~\ref{fig:scaphoid_result}.
The baseline model scored at $74.1\% \pm 1.3\%$ image-wise and $76.4\% \pm 0.9\%
$ study-wise under s-512, and $75.5\% \pm 1.2\%$ image-wise and $79.8\% \pm 2.1\%$ study-wise under s-768 respectively.
Interestingly, $t$-test failed to reject the null hypothesis between the image-wise baseline results ($p = 0.13$) but succeeded between the study-wise results ($p<0.01$), demonstrating the necessity of the view ensemble. For the rest of the scaphoid and ankle fracture experiments, we describe the results by the study-wise accuracy.

\begin{figure*}
    \centering
    \resizebox{\textwidth}{!}{
    \begin{tabular}{cccc|cccc|cccc}
    & & & & & & $g(\mathbf{M}_c)$ & $g(\mathbf{M}'_c)$ & & & $g(\mathbf{M}_c)$ & $g(\mathbf{M}'_c)$ \\
    & & & & 
    \multirow{2}{*}[0.45in]{\shortstack{\includegraphics[width=0.16\textwidth]{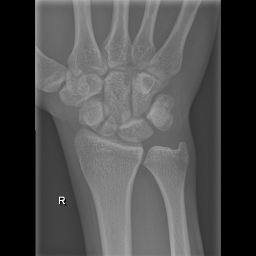}\\No Annotation}} &
    \rotatebox{90}{No Fracture (0.4\%)} &
    \includegraphics[width=0.16\textwidth]{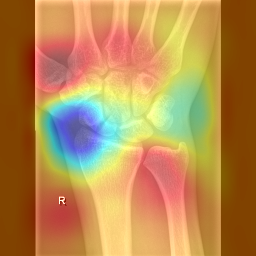} &
    \includegraphics[width=0.16\textwidth]{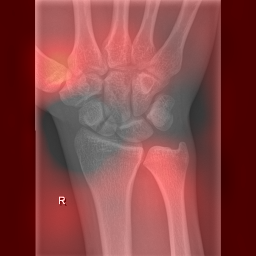} &
    \multirow{2}{*}[0.45in]{\shortstack{\includegraphics[width=0.16\textwidth]{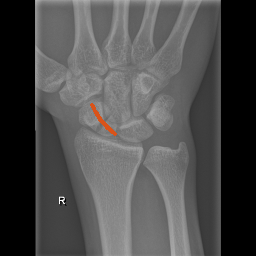}\\Scribble GT}} & 
    \rotatebox{90}{No Fracture (0.3\%)} &
    \includegraphics[width=0.16\textwidth]{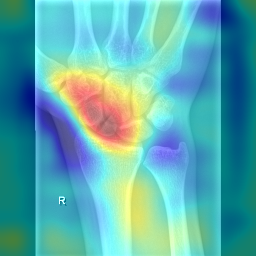} & 
    \includegraphics[width=0.16\textwidth]{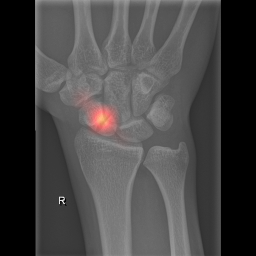} \\
    \multicolumn{4}{c|}{\multirow{2}{*}[0.45in]{\includegraphics[width=0.24\textwidth]{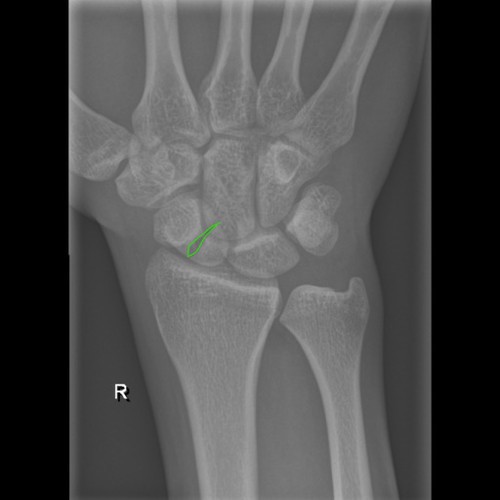}}} &
    &
    \rotatebox{90}{\color{forestgreen}Fracture (99.6\%)\color{black}} &
    \includegraphics[width=0.16\textwidth]{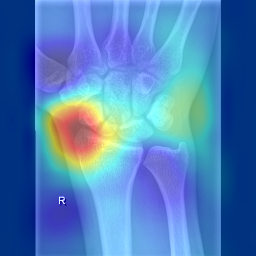} &
    \includegraphics[width=0.16\textwidth]{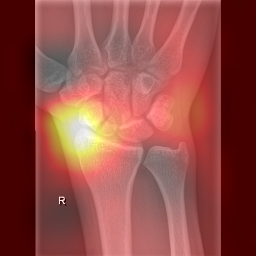} &
    &
    \rotatebox{90}{\color{forestgreen}Fracture (99.7\%)\color{black}} & 
    \includegraphics[width=0.16\textwidth]{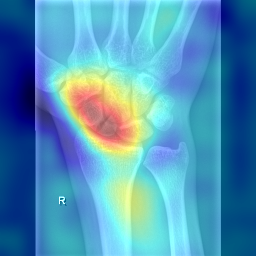} &
    \includegraphics[width=0.16\textwidth]{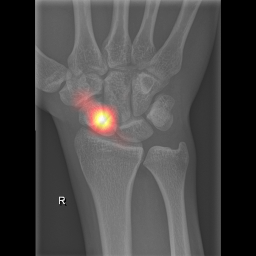} \\
    & & & & \multicolumn{4}{c|}{(b) ResNet Attention without Attention Guidance} & \multicolumn{4}{c}{(c) ResNet Attention Guided by Scribble} \\
    \cline{5-12}
    & & & & & & & & & \\
    & & & & & & $g(\mathbf{M}_c)$ & $g(\mathbf{M}'_c)$ & & & $g(\mathbf{M}_c)$ & $g(\mathbf{M}'_c)$ \\
    \multicolumn{4}{c|}{\shortstack{(a) Original Image with \\ Fracture Indicated}} &
    \multirow{2}{*}[0.45in]{\shortstack{\includegraphics[width=0.16\textwidth]{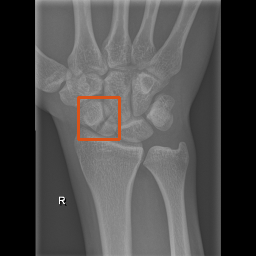}\\BBOX GT}} & 
    \rotatebox{90}{No Fracture (0.6\%)} &
    \includegraphics[width=0.16\textwidth]{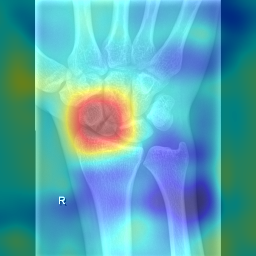} &
    \includegraphics[width=0.16\textwidth]{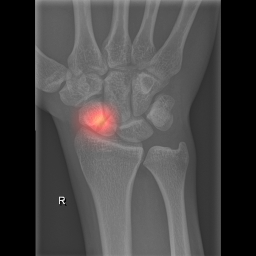} &
    \multirow{2}{*}[0.45in]{\shortstack{\includegraphics[width=0.16\textwidth]{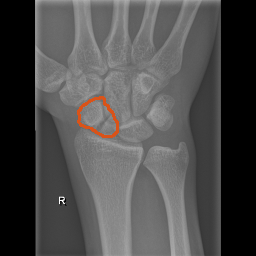}\\Segm. GT}} & 
    \rotatebox{90}{No Fracture (1.0\%)} &
    \includegraphics[width=0.16\textwidth]{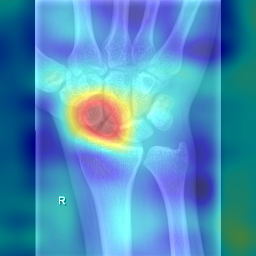} &
    \includegraphics[width=0.16\textwidth]{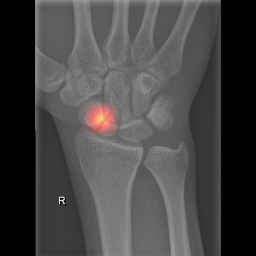} \\
    & & & &
    &
    \rotatebox{90}{\color{forestgreen}Fracture (99.4\%)\color{black}} &
    \includegraphics[width=0.16\textwidth]{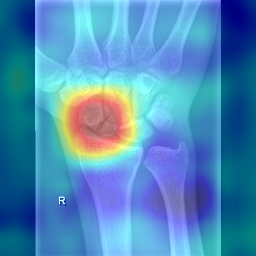} &
    \includegraphics[width=0.16\textwidth]{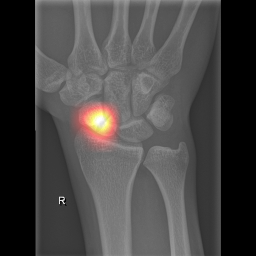} &
    &
    \rotatebox{90}{\color{forestgreen}Fracture (99.0\%)\color{black}} &
    \includegraphics[width=0.16\textwidth]{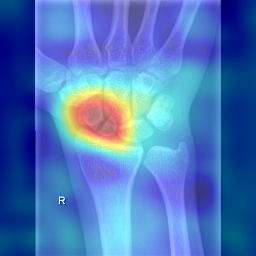} &
    \includegraphics[width=0.16\textwidth]{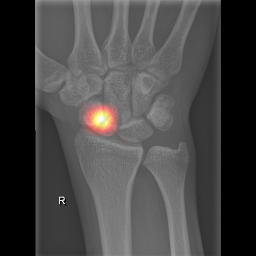} \\
    & & & & 
    \multicolumn{4}{c}{(d) ResNet Attention Guided by Bounding Box} &
    \multicolumn{4}{c}{(e) ResNet Attention Guided by Segmentation}
    \\
    \end{tabular}
    }
    \caption{The visual comparison of the network attention maps for a scaphoid fracture (proximal pole fracture) image. The location of the fracture is marked in green colour in (a), and the predictions in (b) to (e) are all fracture with high confidence. (c) to (e) demonstrate that the proposed attention guidance can effectively guide the network to ground the prediction on the designated area. Note that the colouring of $g(\mathbf{M}_c)$ maps are normalized by min/max($g(\mathbf{M}_c)$) but the $g(\mathbf{M}'_c)$ maps are normalized by min/max($g(\mathbf{M}')$) to correspond the shown intensity with the prediction probabilities. Note that for the vanilla CAM in (b), $g(\mathbf{M}'_c)$ and $g(\mathbf{M}_c)$ are the same and we only impose different colour schemes.}
    \label{fig:scaphoid_example}
\end{figure*}

\begin{figure*}
    \centering
    \resizebox{\textwidth}{!}{
    \begin{tabular}{cccc|cccc|cccc}
    & & & & & & $g(\mathbf{M}_c)$ & $g(\mathbf{M}'_c)$ & & & $g(\mathbf{M}_c)$ & $g(\mathbf{M}'_c)$ \\
    \multicolumn{4}{c|}{\multirow{2}{*}[1.1in]{\includegraphics[width=0.24\textwidth]{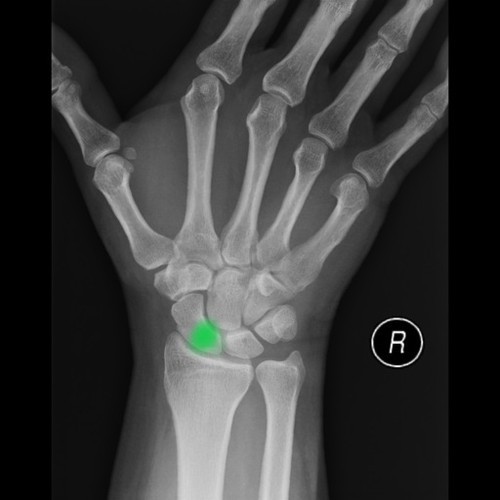}}}
    & 
    \multirow{2}{*}[0.45in]{\shortstack{\includegraphics[width=0.16\textwidth]{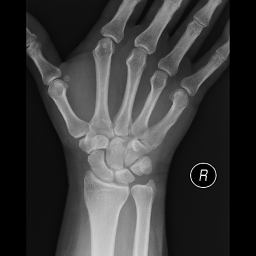}\\No Annotation}} &
    \rotatebox{90}{No Fracture ($<$0.1\%)} &
    \includegraphics[width=0.16\textwidth]{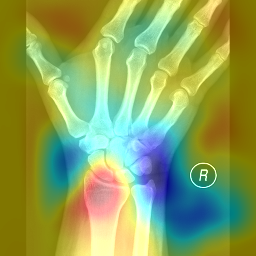} &
    \includegraphics[width=0.16\textwidth]{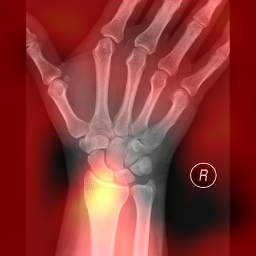} &
    \multirow{2}{*}[0.45in]{\shortstack{\includegraphics[width=0.16\textwidth]{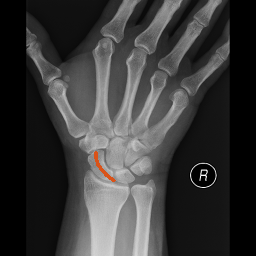}\\Scribble GT}} & 
    \rotatebox{90}{No Fracture ($<$0.1\%)} &
    \includegraphics[width=0.16\textwidth]{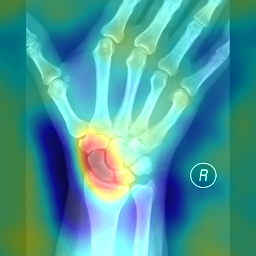} & 
    \includegraphics[width=0.16\textwidth]{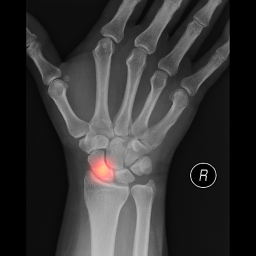} \\
    \multicolumn{4}{c|}{\multirow{2}{*}[0.51in]{\includegraphics[width=0.24\textwidth]{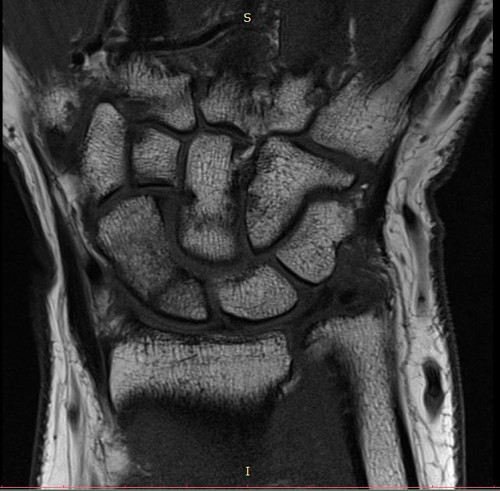}}} &
    &
    \rotatebox{90}{\color{forestgreen}Fracture ($>$99.9\%)\color{black}} &
    \includegraphics[width=0.16\textwidth]{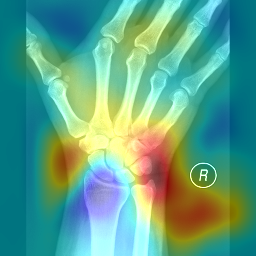} &
    \includegraphics[width=0.16\textwidth]{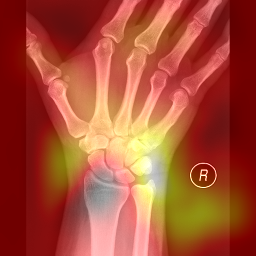} &
    &
    \rotatebox{90}{\color{forestgreen}Fracture ($>$99.9\%)\color{black}} & 
    \includegraphics[width=0.16\textwidth]{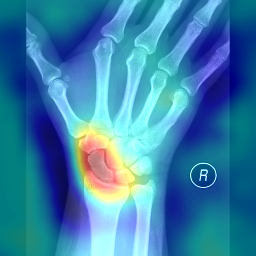} &
    \includegraphics[width=0.16\textwidth]{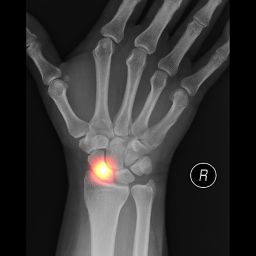} \\
    & & & & \multicolumn{4}{c|}{(b) ResNet Attention without Attention Guidance} & \multicolumn{4}{c}{(c) ResNet Attention Guided by Scribble} \\
    \cline{5-12}
    & & & & & & & & & \\
        \multicolumn{4}{c|}{\multirow{2}{*}[-0.75in]{\includegraphics[width=0.24\textwidth]{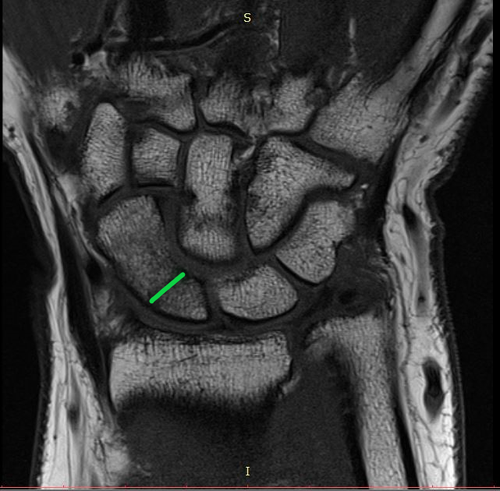}}} & & & $g(\mathbf{M}_c)$ & $g(\mathbf{M}'_c)$ \\
    & & & &
    \multirow{2}{*}[0.45in]{\shortstack{\includegraphics[width=0.16\textwidth]{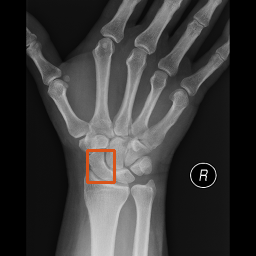}\\BBOX GT}} & 
    \rotatebox{90}{No Fracture (0.1\%)} &
    \includegraphics[width=0.16\textwidth]{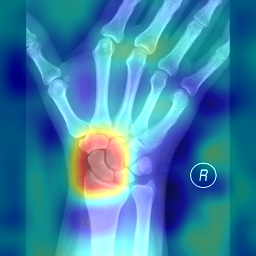} &
    \includegraphics[width=0.16\textwidth]{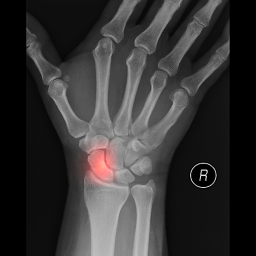} &
    \multirow{2}{*}[0.45in]{\shortstack{\includegraphics[width=0.16\textwidth]{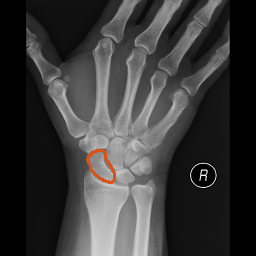}\\Segm. GT}} & 
    \rotatebox{90}{No Fracture (0.5\%)} &
    \includegraphics[width=0.16\textwidth]{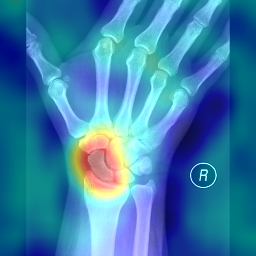} &
    \includegraphics[width=0.16\textwidth]{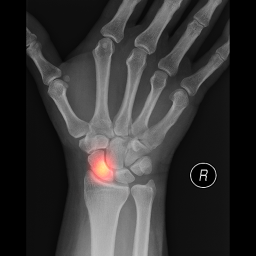} \\
    & & & &
    &
    \rotatebox{90}{\color{forestgreen}Fracture (99.9\%)\color{black}} &
    \includegraphics[width=0.16\textwidth]{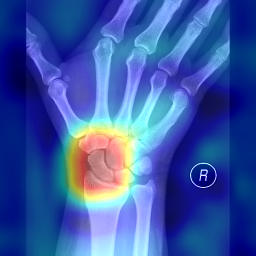} &
    \includegraphics[width=0.16\textwidth]{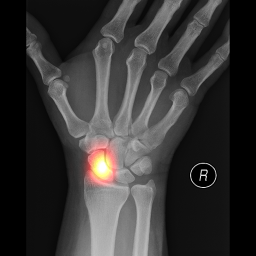} &
    &
    \rotatebox{90}{\color{forestgreen}Fracture (99.5\%)\color{black}} &
    \includegraphics[width=0.16\textwidth]{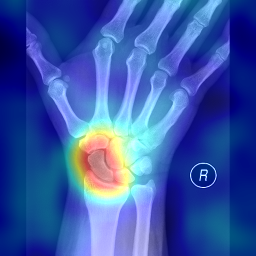} &
    \includegraphics[width=0.16\textwidth]{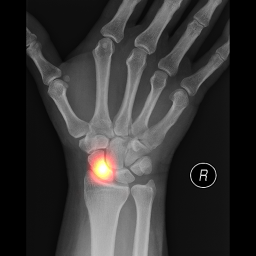} \\
    \multicolumn{4}{c|}{\shortstack{(a) Original Radiograph and \\ Corresponding MR with\\Fracture Region Indicated}} & 
    \multicolumn{4}{c}{(d) ResNet Attention Guided by Bounding Box} &
    \multicolumn{4}{c}{(e) ResNet Attention Guided by Segmentation}
    \\
    \end{tabular}
    }
    \caption{Another visual comparison of the network attention maps for a proximal pole fracture image. The fracture is occult (not visible on any of the radiograph views) but the fracture can be confirmed on MRI as a linear area of low signal (black line) traversing the bone from cortex to cortex on T1-weighted sequence in right-hand side sub-images in (a). In (b) the attention without guidance (\ie, vanilla CAM) focused on the distal radius and bottom part of the scaphoid as evidence of no fracture prediction and on the soft tissue, triquetum and ulnar styloid (tip of the ulna) to determine fracture. In (c) to (e), all three styles of guidance localized more towards the lower region of the scaphoid, which is close to the annotated fracture region indicated by the T1-weighted sequence on MRI in (a).}
    \label{fig:scaphoid_occult_example}
\end{figure*}
\begin{figure*}
    \centering
    \resizebox{\textwidth}{!}{
    \begin{tabular}{cccc|cccc|cccc}
    & & & & & & $g(\mathbf{M}_c)$ & $g(\mathbf{M}'_c)$ & & & $g(\mathbf{M}_c)$ & $g(\mathbf{M}'_c)$ \\
    & & & & 
    \multirow{2}{*}[0.45in]{\shortstack{\includegraphics[width=0.16\textwidth]{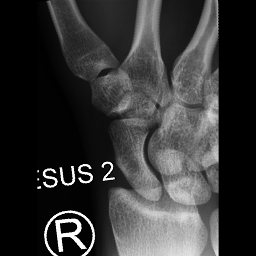}\\No Annotation}} &
    \rotatebox{90}{No Fracture (8.6\%)} &
    \includegraphics[width=0.16\textwidth]{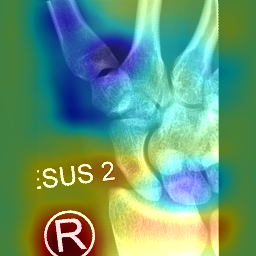} &
    \includegraphics[width=0.16\textwidth]{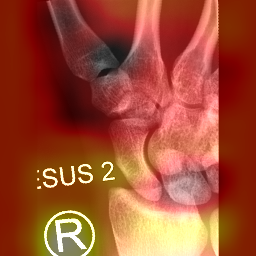} &
    \multirow{2}{*}[0.45in]{\shortstack{\includegraphics[width=0.16\textwidth]{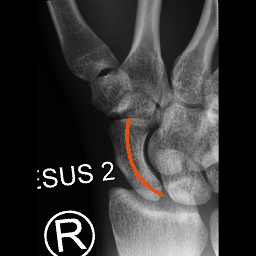}\\Scribble GT}} & 
    \rotatebox{90}{No Fracture (0.8\%)} &
    \includegraphics[width=0.16\textwidth]{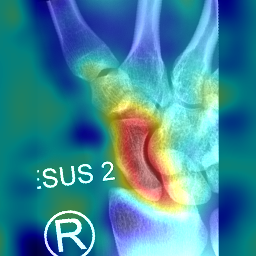} & 
    \includegraphics[width=0.16\textwidth]{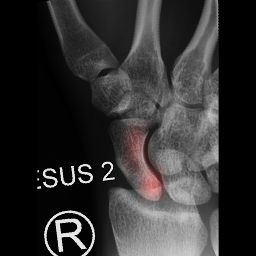} \\
    \multicolumn{4}{c|}{\multirow{2}{*}[0.45in]{\includegraphics[width=0.24\textwidth]{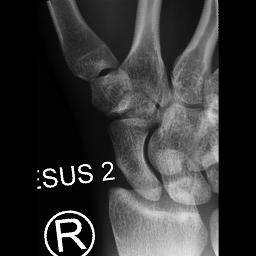}}} &  
    &
    \rotatebox{90}{\color{red}Fracture (91.4\%)\color{black}} &
    \includegraphics[width=0.16\textwidth]{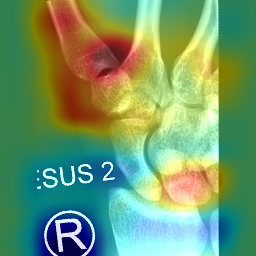} &
    \includegraphics[width=0.16\textwidth]{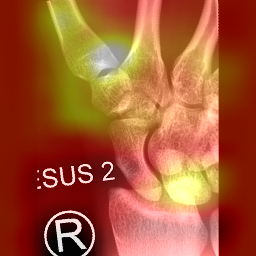} &
    &
    \rotatebox{90}{\color{red}Fracture (99.2\%)\color{black}} & 
    \includegraphics[width=0.16\textwidth]{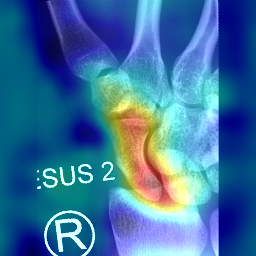} &
    \includegraphics[width=0.16\textwidth]{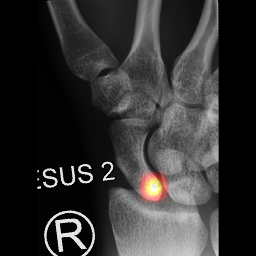} \\
    & & & & \multicolumn{4}{c|}{(b) ResNet Attention without Attention Guidance} & \multicolumn{4}{c}{(c) ResNet Attention Guided by Scribble} \\
    \cline{5-12}
    & & & & & & & & & \\
    & & & & & & $g(\mathbf{M}_c)$ & $g(\mathbf{M}'_c)$ & & & $g(\mathbf{M}_c)$ & $g(\mathbf{M}'_c)$ \\
    \multicolumn{4}{c|}{\shortstack{(a) Original Image with \\No  Fracture}} &
    \multirow{2}{*}[0.45in]{\shortstack{\includegraphics[width=0.16\textwidth]{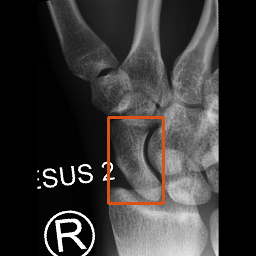}\\BBOX GT}} & 
    \rotatebox{90}{\color{forestgreen}No Fracture (97.7\%)\color{black}} &
    \includegraphics[width=0.16\textwidth]{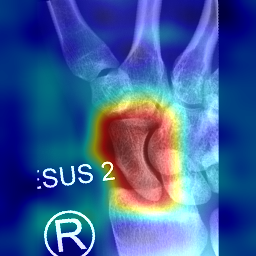} &
    \includegraphics[width=0.16\textwidth]{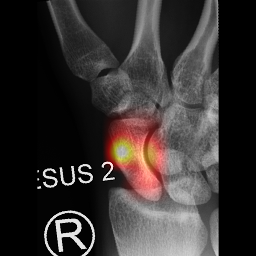} &
    \multirow{2}{*}[0.45in]{\shortstack{\includegraphics[width=0.16\textwidth]{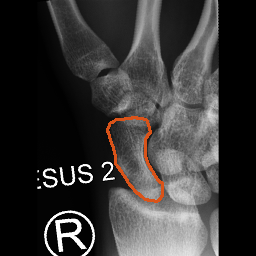}\\Segm. GT}} & 
    \rotatebox{90}{\color{forestgreen}No Fracture (68.4\%)\color{black}} &
    \includegraphics[width=0.16\textwidth]{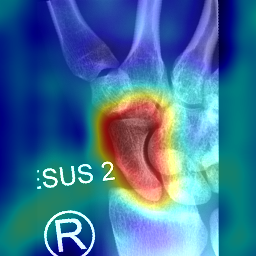} &
    \includegraphics[width=0.16\textwidth]{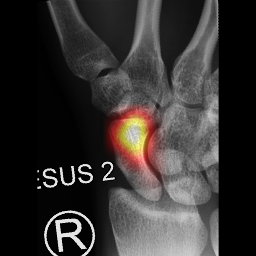} \\
    & & & &
    &
    \rotatebox{90}{Fracture (2.3\%)} &
    \includegraphics[width=0.16\textwidth]{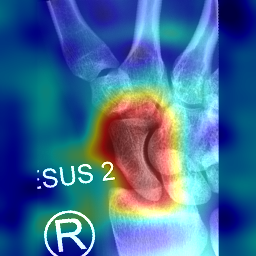} &
    \includegraphics[width=0.16\textwidth]{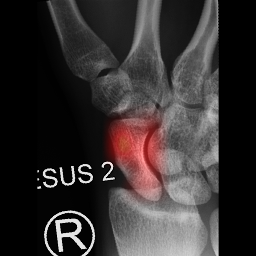} &
    &
    \rotatebox{90}{Fracture (31.6\%)} &
    \includegraphics[width=0.16\textwidth]{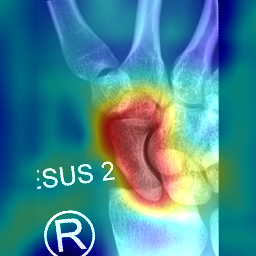} &
    \includegraphics[width=0.16\textwidth]{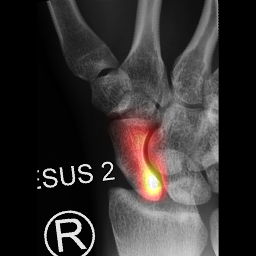} \\
    & & & & 
    \multicolumn{4}{c}{(d) ResNet Attention Guided by Bounding Box} &
    \multicolumn{4}{c}{(e) ResNet Attention Guided by Segmentation}
    \\
    \end{tabular}
    }
    \caption{The visual comparison of the network attention maps for a no fracture image. In (b), the attention without guidance (\ie vanilla CAM) leveraged the salient texts as evidence for the no fracture class and the first carpometacarpal joint and lunate bone for the fracture class. In (c), for the scribble style, the overlapping edge of the radius with the proximal pole of the scaphoid was discovered as a fracture. In (d), the bounding box guidance correctly predicted the image as no fracture, using the waist of the scaphoid as evidence for no fracture and also ignoring the projection of the overlapping radius edge. Finally in (e) for the segmentation style, the distal pole of the scaphoid triggered a high activation for no fracture while the overlapping projection of the radius triggered fracture.}
    \label{fig:scaphoid_failure_example_190_0004_0001}
\end{figure*}

The L1 and L2-regularization penalize on the magnitude of $\mathbf{M}''_c$ from Eq. (\ref{eq:up_sampled_cam}), which replace the attention guidance loss, i.e., the second term in Eq.~(\ref{eq:loss}). $\lambda$ controls the influence of the regularization, and at 1e-8 we observe their highest scores $80.0\% \pm 2.4\%$ on s-512 using L1 (a significant 3.6\% mean accuracy improvement, $p = 0.02$) and $82.0\% \pm 2.5\%$ on s-768 using L2 (a marginal 2.2\% improvement, $p = 0.16$).

For a quantitative comparison to the existing attention mechanism methods, we adapt the official implementation of three methods, \ie, residual attention~\cite{wang2017residual}, self attention~\cite{zhang2019self}, and pyramid attention~\cite{li2018pyramid}, where the commonality between the three methods is that they were all designed to be placed as an additional processing module after $\mathbf{F}$ and the processed tensor shares the same size as $\mathbf{F}$.
As an overview, we found that the explored attention mechanisms can improve over the baseline (\ie, no attention is used), however most of them are of marginal improvement except that of self attention, which achieves significant improvement ($p = 0.02$) by a mean 2.8\% accuracy improvement to $79.2\% \pm 1.8\%$ on s-512.

For the proposed attention guided networks, the best performance achieved using the scribble focus region on s-512 is with $\lambda=1$, resulting in a 6.8\% improvement over the baseline; for s-768, the best performance lands at $\lambda=0.01$ with a 5.2\% improvement.
The improvement using bounding box and segmentation style guidance are in similar quantity, which is 7.8\% on s-512 and 4.6\% on s-768 for bounding box and 6.6\% on s-512 and 4.2\% on s-768 for segmentation respectively.
The main conclusions of the experiments are the followings. 
\begin{itemize}
    \item The explored L1/L2 regularization and attention mechanisms only perform marginally better than the baseline does, where the level of improvement is similar to that of the proposed attention guidance with $\lambda = 0$, \ie, enabling the attention to be formed only via the task training signal or normalized based on its norm but not receiving the human guidance.
    \item The usefulness of the attention guidance is subject to an appropriate influence (controllable by $\lambda$) during the CNN training. %
    \item The scribble, bounding box, and segmentation styles result in similar performance. 
    This makes the scribble style focus region more favorable as we have investigated the annotating speed, which is 24.1s per study for the scribble style, 31.1s for the bounding box, and 116.5s  for the segmentation style on average, computed based on annotating 10 studies each.
    \item The best performance from each input size, \ie, bounding box at $\lambda=1$ on s-512 ( $84.2\% \pm 0.8\%$) and scribble at $\lambda=0.01$ on s-768 ($85.0\% \pm 1.4\%$) are not significantly different from each other ($p = 0.31$), making s-768 less favourable as it costs a much larger amount of computation overhead (1hr training time, 10.6GB GPU memory usage) versus s-512 (0.5hr, 9.9GB).
\end{itemize} 
Finally, in the work by Langerhuizen \etal~\cite{langerhuizen2020deep}, the reported study-wise accuracy is 72\% using a VGG-16~\cite{simonyan2014very} trained by manually cropped image containing the scaphoid. 
In comparison, the proposed attention guidance method reduces the manual cropping effort at the inference stage and the performance under the optimal configuration is on par with the 84\% human performance computed by five orthopaedic surgeons as reported in~\cite{langerhuizen2020deep}.

{\bf Visual Comparison:}
in Fig.~\ref{fig:scaphoid_example}, we show a qualitative comparison of network attention with and without the proposed attention guidance.
In Fig.~\ref{fig:scaphoid_example}-(b), the no fracture class $\mathbf{M}_c$ attention from the baseline ResNet hovers on the surrounding bones including the base of metacarpals I-IV, the ulnar head and on the ``R'' radiograph tag (used to indicate the image is of a right hand).
In addition, the CNN also has a negative attention (region with negative attention values, shown in the blue color, can be thought as the network using the existence of certain patterns as the negative evidence, \ie, cat does not have beak) over the scaphoid bone but the centre sits on the radial styloid region.
The fracture class attention is complementary to the no fracture attention.
This case is exemplary such that the baseline ResNet could produce a very confident (and correct) prediction, but the grounding location of such decision is not relevant to the scaphoid fracture. 
In Fig.~\ref{fig:scaphoid_example}-(c) to (e), we show that with any of the three attention guidance types, $\mathbf{M}_c$ attention is appropriately positioned on the scaphoid bone and its surrounding area, and $\mathbf{M}'_c$ correctly discovers the actual proximal pole fracture and suppresses the non-relevant positive (yellow) and negative (blue) attention regions of $\mathbf{M}_c$.
In addition to this example, we present an occult scaphoid fracture example in Fig.~\ref{fig:scaphoid_occult_example}, and an example in Fig.~\ref{fig:scaphoid_failure_example_190_0004_0001} displaying the no fracture case that was misclassified by either no attention or the scribble guidance. 
In Fig.~\ref{fig:scaphoid_failure_example_190_0004_0001}, both the scribble and segmentation guidance detect the overlapping border of the radius onto the scaphoid bone as fracture, although with the segmentation guidance (in contrast to the scribble guidance) this is outweighed by the no fracture activation in the distal pole of the scaphoid resulting in a correct prediction.

\begin{figure}[!htbp]
    \centering
    \resizebox{0.9\columnwidth}{!}{
    \includegraphics[width=0.5\textwidth]{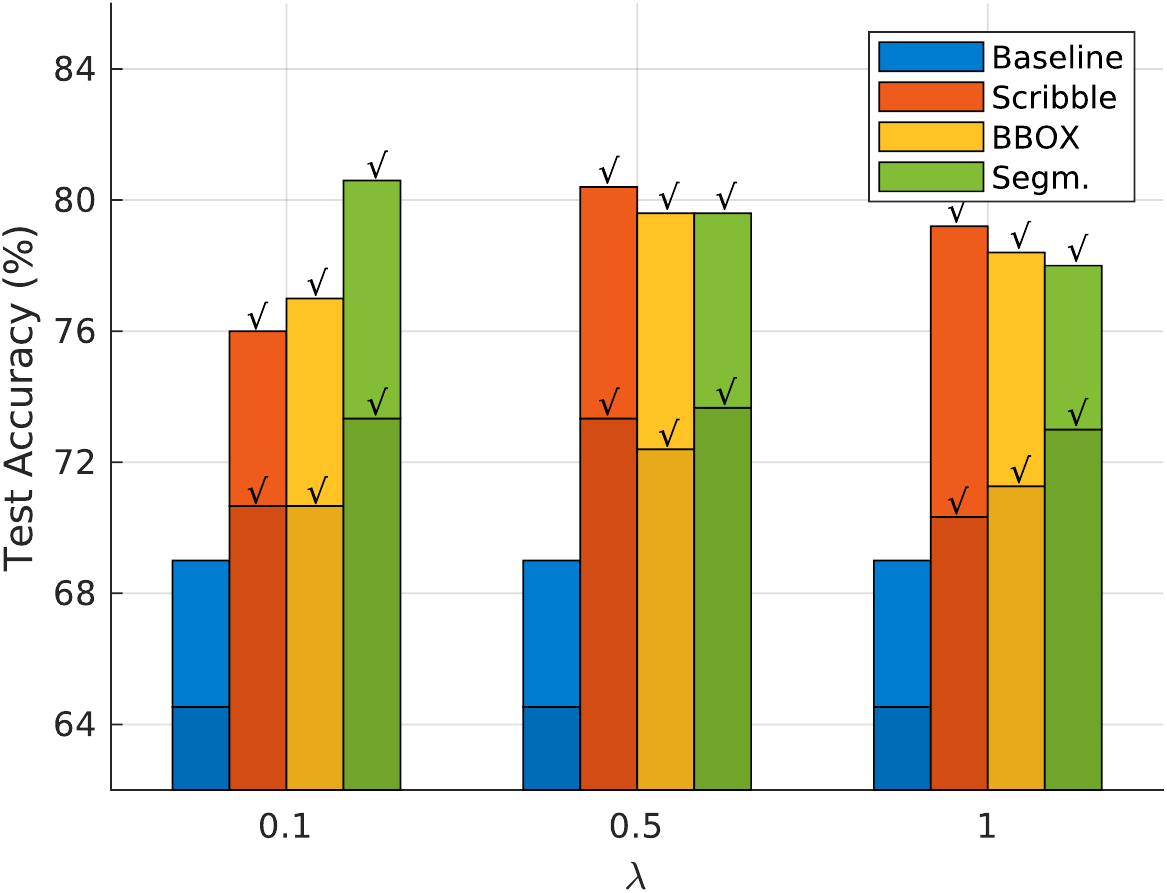}
    }
    \caption{The result comparison on the ankle fracture dataset as a function of the $\lambda$ assigned to the proposed attention regularization loss in Eq.~(\ref{eq:loss}), supervised using scribble, bounding box (BBOX), or segmentation (Segm.) style focus region.}
    \label{fig:ankle_result}
\end{figure}

\begin{table*}[!htbp]
    \caption{The ankle fracture dataset experiment results on the test set and the external test set in classification accuracy (\%). Each table cell shows the statistical result computed by five repetitive runs of the same setting. The highest classification performance is highlighted for each column.}
    \centering
    \resizebox{\textwidth}{!}{
    \begin{tabular}{|c|c|c|c|c|c|c|c|c|c|c|c|c|}
    \hline
    \multirow{2}{*}{Methods}	&	\multicolumn{6}{c|}{Test Set}													&	\multicolumn{6}{c|}{External Test Set}												\\ \cline{2-13}
	&			\multicolumn{3}{c|}{Image-wise}			&			\multicolumn{3}{c|}{Study-wise}					&			\multicolumn{3}{c|}{Image-wise}			&			\multicolumn{3}{c|}{Study-wise}				\\ \hline
Baseline	& \multicolumn{3}{c|}{$			64.5 \pm 1.7			$} & \multicolumn{3}{c|}{$			69.0 \pm 2.8			$} 		& \multicolumn{3}{c|}{$			68.7 \pm 2.7			$} & \multicolumn{3}{c|}{$			75.4 \pm 2.5			$} 	 \\ \hline
L1 Reg. ($\lambda$=1e-9)	& \multicolumn{3}{c|}{$			62.3 \pm 9.5			$} & \multicolumn{3}{c|}{$			67.0 \pm 11.9			$} 		& \multicolumn{3}{c|}{$			70.2 \pm 6.0			$} & \multicolumn{3}{c|}{$			78.1 \pm 7.6			$} 	 \\ \hline
L2 Reg. ($\lambda$=1e-8)	& \multicolumn{3}{c|}{$			67.0 \pm 5.5			$} & \multicolumn{3}{c|}{$			72.8 \pm 6.7			$} 		& \multicolumn{3}{c|}{$			74.5 \pm 2.0			$} & \multicolumn{3}{c|}{$			81.7 \pm 4.3			$} 	 \\ \hline
Residual Attn.~\cite{wang2017residual}	& \multicolumn{3}{c|}{$			62.5 \pm 1.6			$} & \multicolumn{3}{c|}{$			65.0 \pm 1.2			$} 		& \multicolumn{3}{c|}{$			68.7 \pm 1.4			$} & \multicolumn{3}{c|}{$			75.5 \pm 1.0			$} 	 \\ \hline
Self Attn.~\cite{zhang2019self}	& \multicolumn{3}{c|}{$			68.9 \pm 2.5			$} & \multicolumn{3}{c|}{$			74.6 \pm 3.1			$} 		& \multicolumn{3}{c|}{$			70.4 \pm 2.0			$} & \multicolumn{3}{c|}{$			78.6 \pm 1.8			$} 	 \\ \hline
Pyramid Attn.~\cite{li2018pyramid}	& \multicolumn{3}{c|}{$			55.4 \pm 6.2			$} & \multicolumn{3}{c|}{$			59.4 \pm 9.6			$} 		& \multicolumn{3}{c|}{$			61.5 \pm 7.2			$} & \multicolumn{3}{c|}{$			67.6 \pm 7.8			$} 	 \\ \hline
Attn. Guidance	&	Scribble	&	BBOX	&	Segm.	&	Scribble	&	BBOX	&	Segm.			&	Scribble	&	BBOX	&	Segm.	&	Scribble	&	BBOX	&	Segm.		\\ \hline
$\lambda$ = 0	& \multicolumn{3}{c|}{$			65.2 \pm 4.0			$} & \multicolumn{3}{c|}{$			70.4 \pm 5.9			$} 		& \multicolumn{3}{c|}{$			73.5 \pm 4.1			$} & \multicolumn{3}{c|}{$			80.5 \pm 4.6			$} 	 \\ \hline
$\lambda$ = 0.1	& $	70.7 \pm 1.6	$ & $	70.7 \pm 1.8	$ & $	73.3 \pm 2.8	$ & $	76.0 \pm 2.2	$ & $	77.0 \pm 3.7	$ & $	\bf 80.6 \pm 3.8		$	& $	\bf 79.6 \pm 1.2	$ & $	77.5 \pm 3.5	$ & $	\bf 79.8 \pm 1.6	$ & $	86.4 \pm 1.9	$ & $	85.6 \pm 1.9	$ & $	\bf 86.4 \pm 2.4		$ \\ \hline
$\lambda$ = 0.5	& $	\bf 73.3 \pm 2.6	$ & $	\bf 72.4 \pm 2.2	$ & $	\bf 73.7 \pm 2.8	$ & $	\bf 80.4 \pm 3.9	$ & $	\bf 79.6 \pm 3.2	$ & $	79.6 \pm 2.0		$	& $	79.4 \pm 2.5	$ & $	\bf 79.3 \pm 0.9	$ & $	79.6 \pm 1.2	$ & $	\bf 87.0 \pm 2.1	$ & $	\bf 86.6 \pm 1.7	$ & $	86.1 \pm 1.1		$ \\ \hline
$\lambda$ = 1	& $	70.3 \pm 3.3	$ & $	71.3 \pm 2.1	$ & $	73.0 \pm 1.6	$ & $	79.2 \pm 4.0	$ & $	78.4 \pm 3.1	$ & $	78.0 \pm 1.6		$	& $	77.1 \pm 2.1	$ & $	78.7 \pm 1.7	$ & $	80.1 \pm 2.2	$ & $	84.5 \pm 1.5	$ & $	84.6 \pm 1.4	$ & $	86.0 \pm 1.6		$

    \\ \hline
    \end{tabular}
    }
    \label{tab:ankle_table}
\end{table*}

\subsection{Ankle Fracture Classification}
\label{sec:ankle_exp}

{\bf Experiment setting and network specification:}
for the ankle fracture classification dataset, the network specification and training parameters are identical to the ones for the scaphoid fracture dataset except explicit sample importance re-weighting during training to avoid biases towards the high population classes.
In this experiment, we compare the performance also using scribble, bounding box, and segmentation style focus regions.

\begin{figure*}
    \centering
    \resizebox{\textwidth}{!}{
    \begin{tabular}{c|ccccccc}
    \multirow{4}{*}[-0.325in]{
    \includegraphics[width=0.24\textwidth]{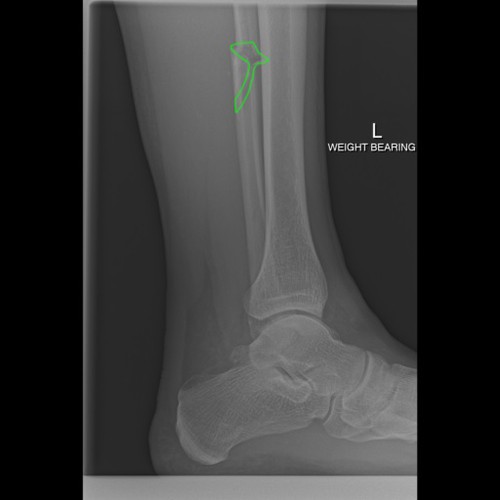}}
    & & & & Weber A & Weber B & Weber C & No Fracture \\
    & &
    \includegraphics[width=0.13\textwidth]{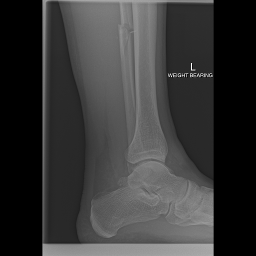} &
    &
    \includegraphics[width=0.13\textwidth]{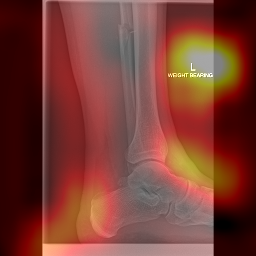} &
    \includegraphics[width=0.13\textwidth]{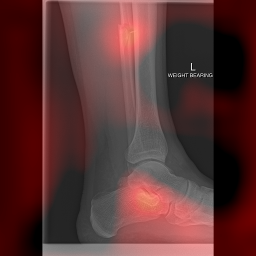} &
    \includegraphics[width=0.13\textwidth]{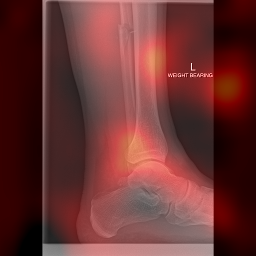} &
    \includegraphics[width=0.13\textwidth]{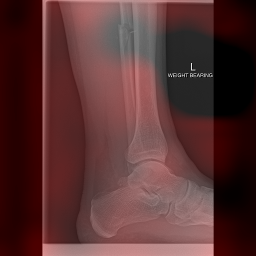} \\
    & & No Annotation & & \color{red}81.7\%\color{black} & 2.5\% & 13.4\% & 2.4\% \\
    & &
    \includegraphics[width=0.13\textwidth]{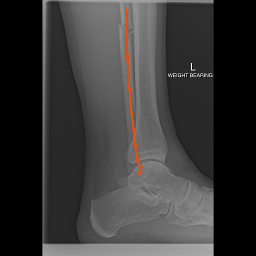} &
    &
    \includegraphics[width=0.13\textwidth]{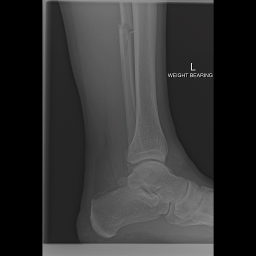} &
    \includegraphics[width=0.13\textwidth]{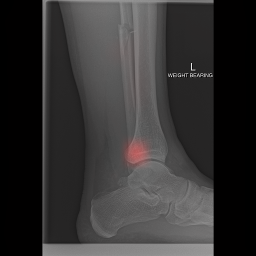} &
    \includegraphics[width=0.13\textwidth]{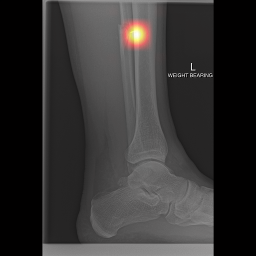} &
    \includegraphics[width=0.13\textwidth]{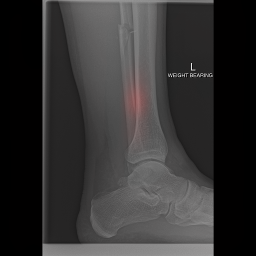} \\
    (a) Original Image with Fracture Indicated & & Scribble & & 1.8\% & 14.5\% & \color{forestgreen}72.9\%\color{black} & 10.8\% \\
    & &
    \includegraphics[width=0.13\textwidth]{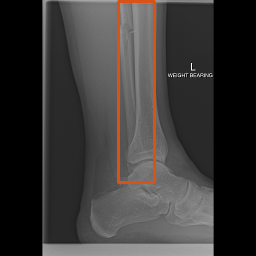} &
    &
    \includegraphics[width=0.13\textwidth]{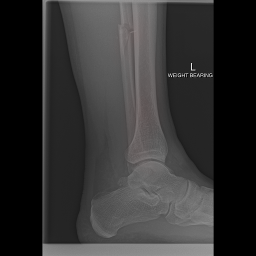} &
    \includegraphics[width=0.13\textwidth]{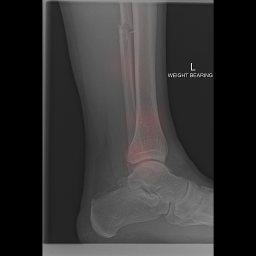} &
    \includegraphics[width=0.13\textwidth]{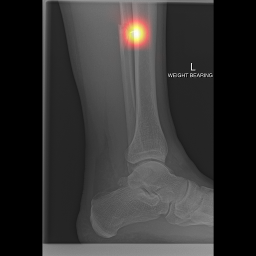} &
    \includegraphics[width=0.13\textwidth]{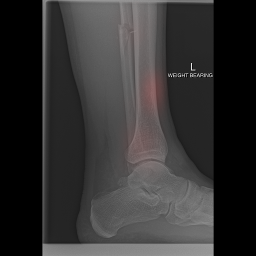} \\
    & & BBOX GT & & 1.1\% & 7.3\% & \color{forestgreen}84.7\%\color{black} & 7.0\% \\
    \multirow{4}{*}[1.625in]{\includegraphics[width=0.24\textwidth]{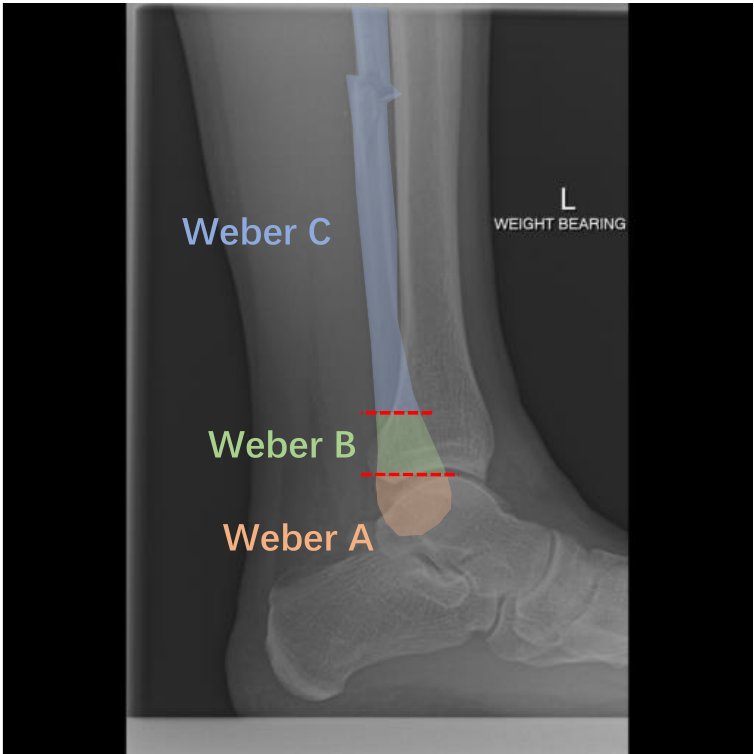}} &
    & 
    \includegraphics[width=0.13\textwidth]{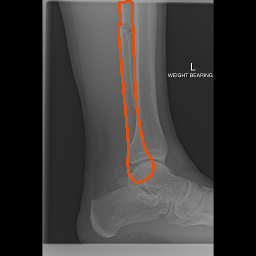} &
    &
    \includegraphics[width=0.13\textwidth]{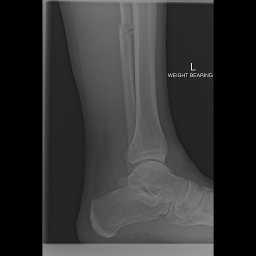} &
    \includegraphics[width=0.13\textwidth]{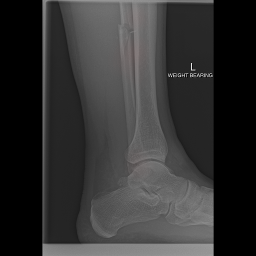} &
    \includegraphics[width=0.13\textwidth]{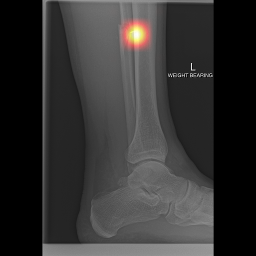} &
    \includegraphics[width=0.13\textwidth]{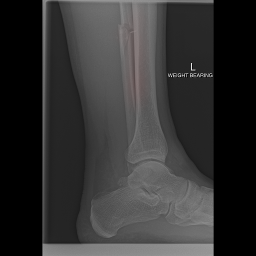} \\
    & & Segm. GT & & 0.2\% & 1.2\% & \color{forestgreen}96.8\%\color{black} & 1.7\% \\
    (b) Weber Classification Visual Representation & \multicolumn{7}{c}{(c) ResNet Attention ($g(\mathbf{M}'_c)$) with or without Attention Guidance} \\
    \end{tabular}
    }
    \caption{The visual comparison of the $g(\mathbf{M}'_c)$ attention maps for a Weber type C fracture image in the lateral view. The location of the fracture is marked in green colour in (a), and an illustration of Weber classification regions in roughly sketched in (b). (c) demonstrates the network attention difference between the ResNets trained with attention guidance or without (\ie, vanilla CAM).}
    \label{fig:fracture_example}
\end{figure*}
\begin{figure*}
    \centering
    \resizebox{\textwidth}{!}{
    \begin{tabular}{c|ccccccc}
    \multirow{4}{*}[-0.325in]{
    \includegraphics[width=0.24\textwidth]{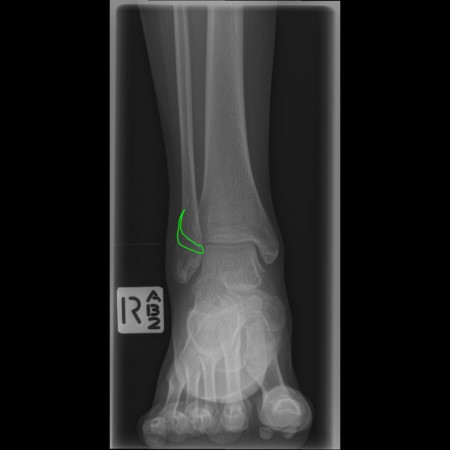}}
    & & & & Weber A & Weber B & Weber C & No Fracture \\
    & &
    \includegraphics[width=0.13\textwidth]{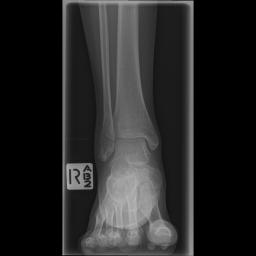} &
    &
    \includegraphics[width=0.13\textwidth]{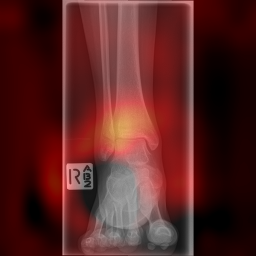} &
    \includegraphics[width=0.13\textwidth]{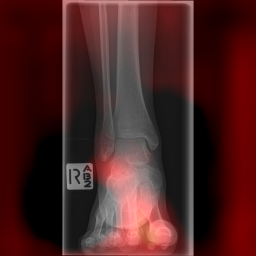} &
    \includegraphics[width=0.13\textwidth]{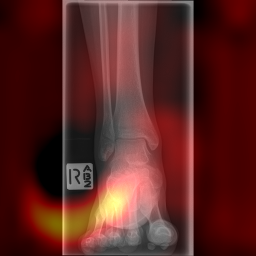} &
    \includegraphics[width=0.13\textwidth]{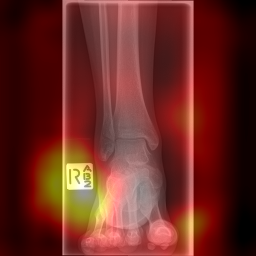} \\
    & & No Annotation & & 13.8\% & 4.1\% & 17.6\% & \color{red}64.5\%\color{black} \\
    & &
    \includegraphics[width=0.13\textwidth]{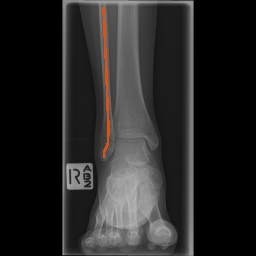} &
    &
    \includegraphics[width=0.13\textwidth]{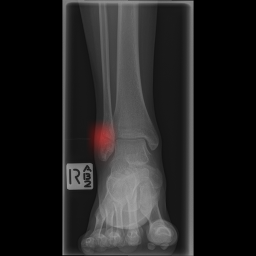} &
    \includegraphics[width=0.13\textwidth]{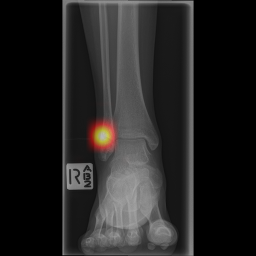} &
    \includegraphics[width=0.13\textwidth]{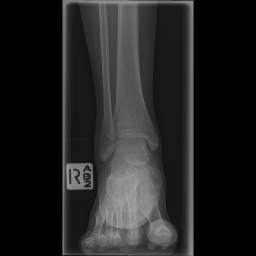} &
    \includegraphics[width=0.13\textwidth]{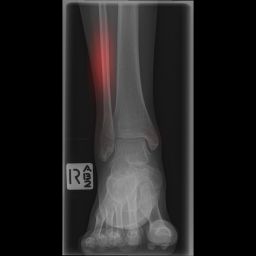} \\
    (a) Original Image with Fracture Indicated & & Scribble & & 4.8\% & \color{forestgreen}58.4\%\color{black} & 0.5\% & 36.3\% \\
    & &
    \includegraphics[width=0.13\textwidth]{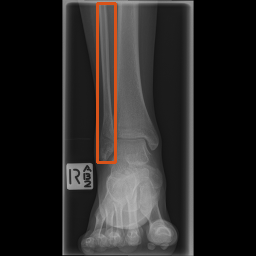} &
    &
    \includegraphics[width=0.13\textwidth]{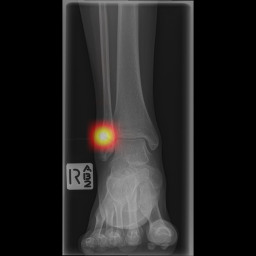} &
    \includegraphics[width=0.13\textwidth]{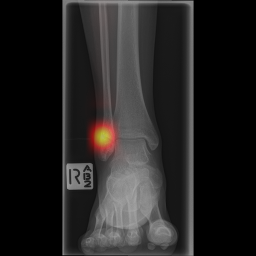} &
    \includegraphics[width=0.13\textwidth]{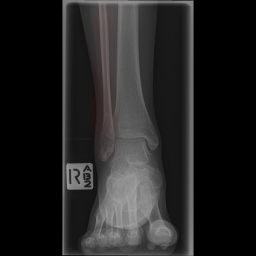} &
    \includegraphics[width=0.13\textwidth]{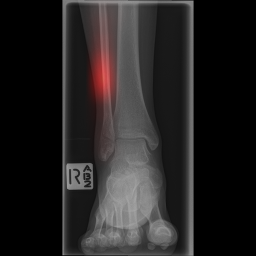} \\
    & & BBOX GT & & 26.0\% & 30.5\% & 1.8\% & \color{red}41.7\%\color{black} \\
    \multirow{4}{*}[1.625in]{\includegraphics[width=0.24\textwidth]{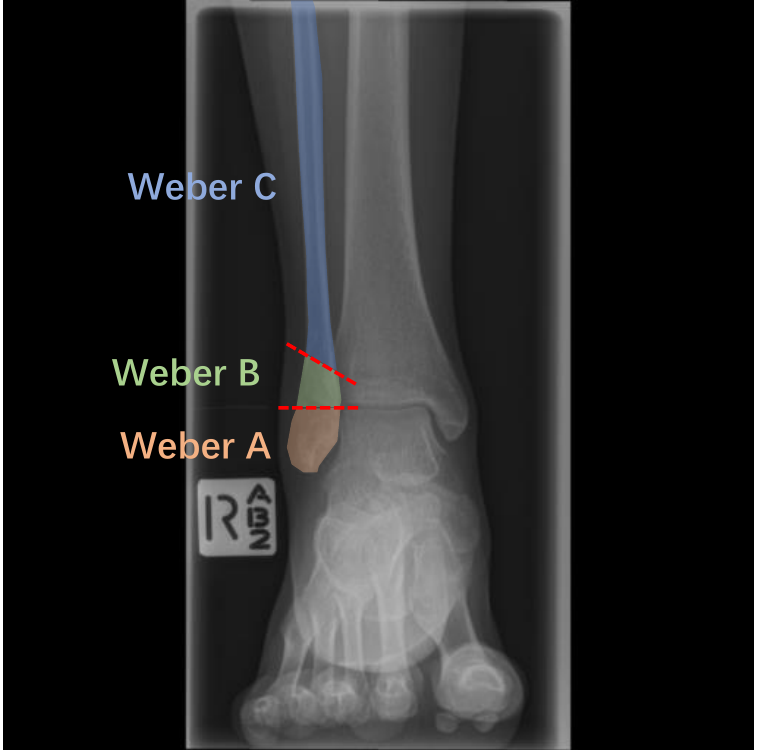}} &
    & 
    \includegraphics[width=0.13\textwidth]{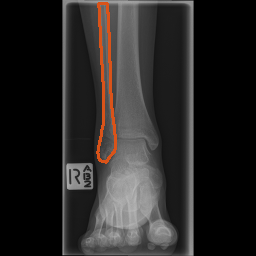} &
    &
    \includegraphics[width=0.13\textwidth]{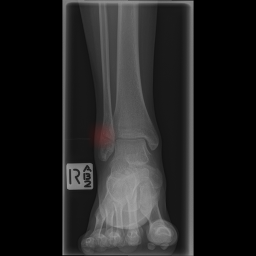} &
    \includegraphics[width=0.13\textwidth]{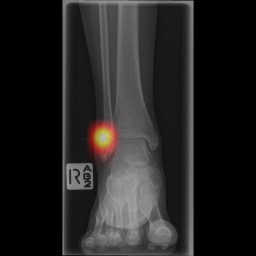} &
    \includegraphics[width=0.13\textwidth]{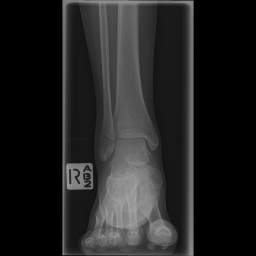} &
    \includegraphics[width=0.13\textwidth]{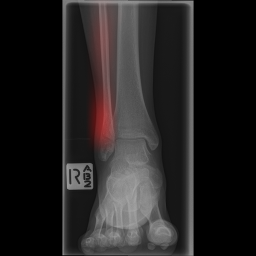} \\
    & & Segm. GT & & 1.2\% & 46.5\% & 0.3\% & \color{red}51.9\%\color{black} \\
    (b) Weber Classification Visual Representation & \multicolumn{7}{c}{(c) ResNet Attention ($g(\mathbf{M}'_c)$) with or without Attention Guidance} \\
    \end{tabular}
    }
    \caption{The visual comparison of the $g(\mathbf{M}'_c)$ attention maps for an exemplary Weber type B fracture image in the AP view. The location of the fracture is marked in green colour in (a), and an illustration of Weber classification regions is in (b). (c) demonstrates the network attention difference between the ResNets trained with attention guidance or without (\ie, vanilla CAM).
    Except for the scribble guidance, the other three examples produced a wrong answer. 
    When trained with no attention, the attention of all three Weber classes failed to track on the exact fracture region.
    When trained with attention guidance, the fracture region is correctly detected but the predictions using bounding box and segmentation are lost to the no fracture class as the no fracture class exceeds the Weber B class using a larger region mild activation. %
    }
    \label{fig:fracture_ap_example}
\end{figure*}
\begin{figure*}
    \centering
    \resizebox{\textwidth}{!}{
    \begin{tabular}{c|ccccccc}
    \multirow{4}{*}[-0.325in]{
    \includegraphics[width=0.24\textwidth]{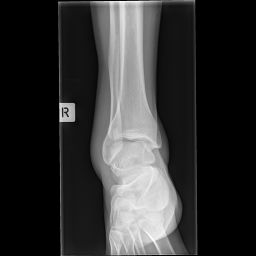}}
    & & & & Weber A & Weber B & Weber C & No Fracture \\
    & &
    \includegraphics[width=0.13\textwidth]{example_jbhi/fracture/no_weak/test/9918_original_l3.png} &
    &
    \includegraphics[width=0.13\textwidth]{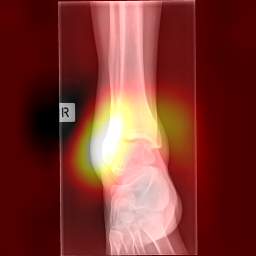} &
    \includegraphics[width=0.13\textwidth]{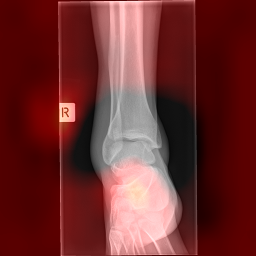} &
    \includegraphics[width=0.13\textwidth]{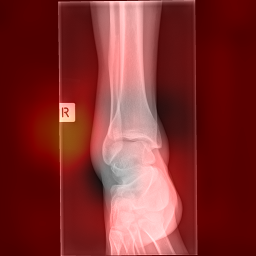} &
    \includegraphics[width=0.13\textwidth]{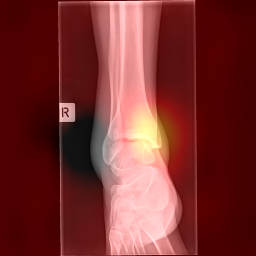} \\
    & & No Annotation & & \color{red}61.3\%\color{black} & 0.7\% & 29.8\% & 8.3\% \\
    & &
    \includegraphics[width=0.13\textwidth]{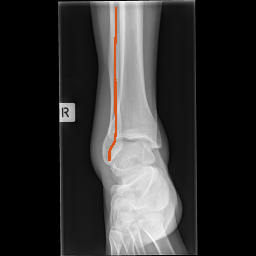} &
    &
    \includegraphics[width=0.13\textwidth]{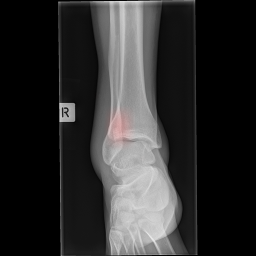} &
    \includegraphics[width=0.13\textwidth]{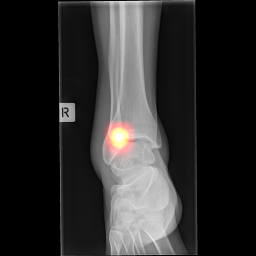} &
    \includegraphics[width=0.13\textwidth]{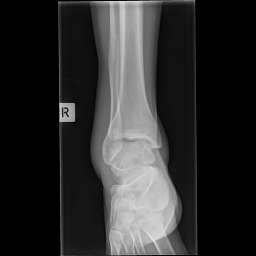} &
    \includegraphics[width=0.13\textwidth]{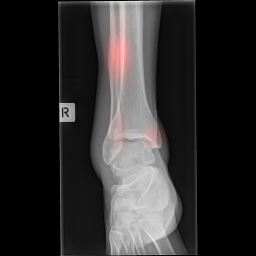} \\
    (a) Original Image with No Fracture & & Scribble & & 4.0\% & \color{red}56.0\%\color{black} & 0.9\% & 13.0\% \\
    & &
    \includegraphics[width=0.13\textwidth]{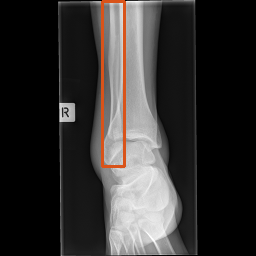} &
    &
    \includegraphics[width=0.13\textwidth]{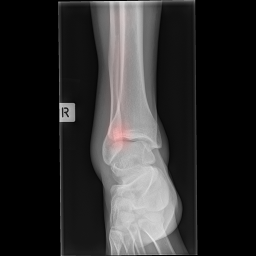} &
    \includegraphics[width=0.13\textwidth]{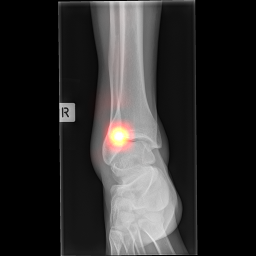} &
    \includegraphics[width=0.13\textwidth]{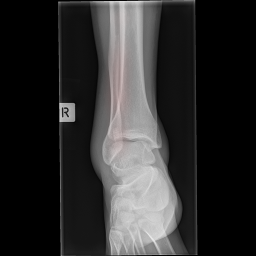} &
    \includegraphics[width=0.13\textwidth]{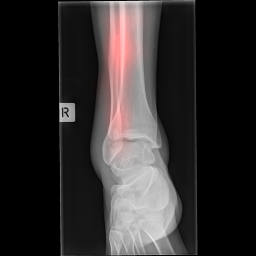} \\
    & & BBOX GT & & 1.5\% & 12.3\% & 0.9\% & \color{forestgreen}85.3\%\color{black} \\
    \multirow{4}{*}[1.625in]{\includegraphics[width=0.24\textwidth]{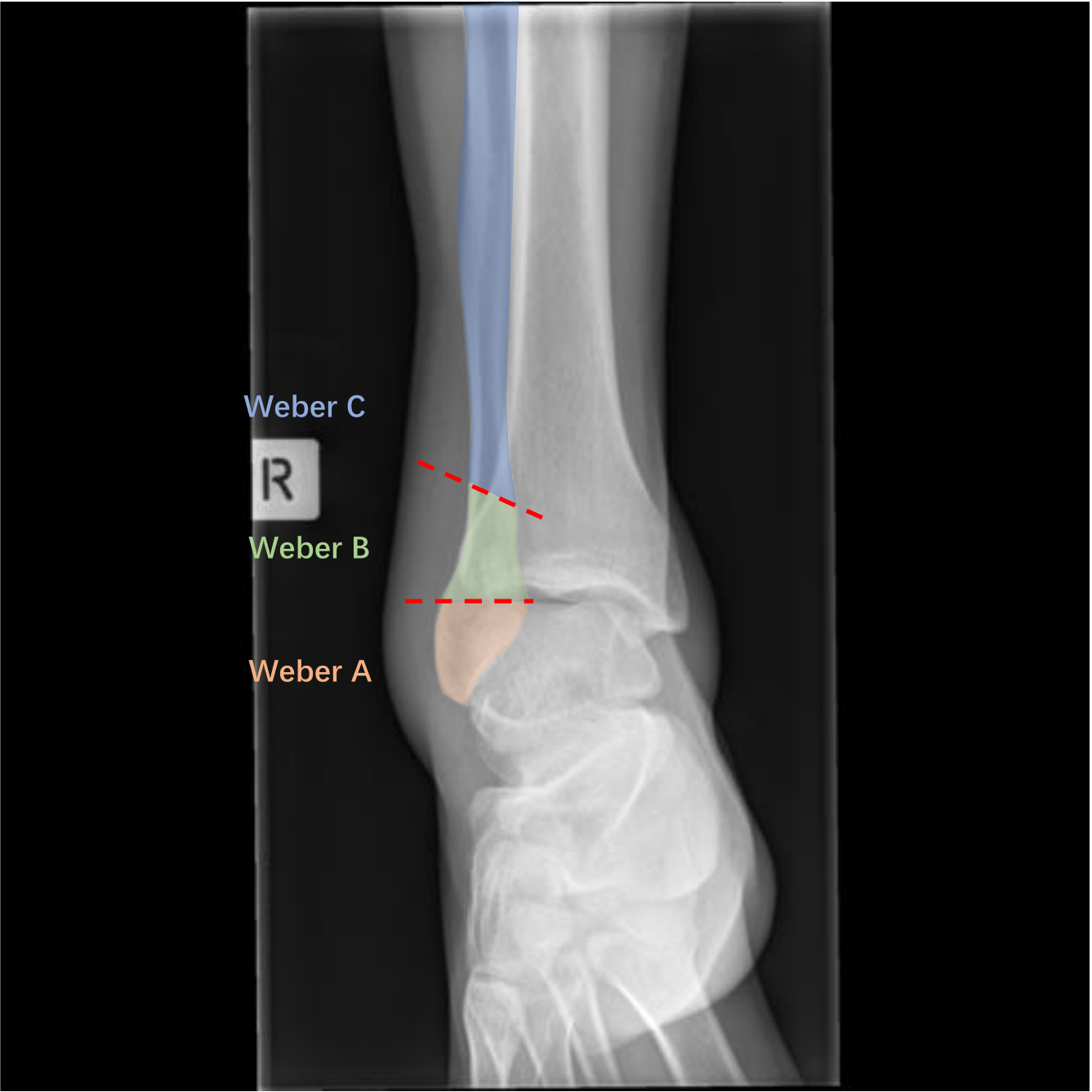}} &
    & 
    \includegraphics[width=0.13\textwidth]{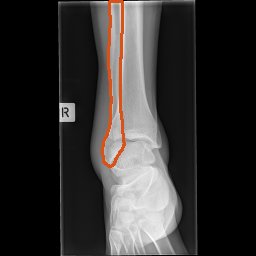} &
    &
    \includegraphics[width=0.13\textwidth]{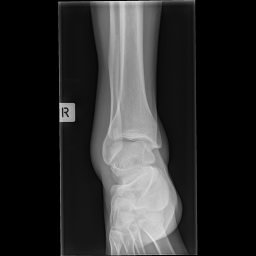} &
    \includegraphics[width=0.13\textwidth]{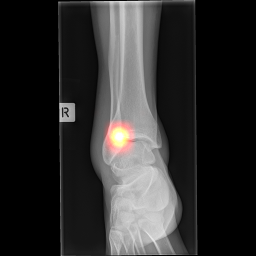} &
    \includegraphics[width=0.13\textwidth]{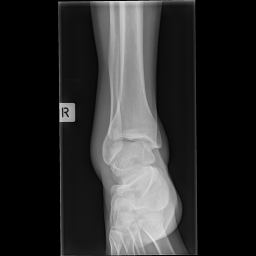} &
    \includegraphics[width=0.13\textwidth]{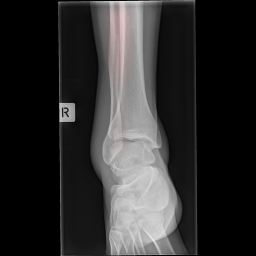} \\
    & & Segm. GT & & 0.3\% & \color{red}95.7\%\color{black} & 0.2\% & 3.8\% \\
    (b) Weber Classification Visual Representation & \multicolumn{7}{c}{(c) ResNet Attention ($g(\mathbf{M}'_c)$) with or without Attention Guidance} \\
    \end{tabular}
    }
    \caption{The visual comparison of the $g(\mathbf{M}'_c)$ attention maps for a no fracture case in the mortise view. 
    The Weber classification regions are illustrated in (b). 
    (c) demonstrates the network attention difference between the ResNets trained with attention guidance or without (\ie, vanilla CAM).
    Without the attention guidance, the network attention activation on lateral malleolus for Weber A, outweighs the activation medial malleolus (on tibia) for No fracture and produces the wrong classification. For the attention guided settings, the attentions in the Weber B region misinterprets the projection of the fibula over the tibia as a fracture, except when bounding box style guidance is used, the mild but large size activation outweighs the Weber B attention activation. We show this case as an example of error. 
    }
    \label{fig:fracture_9918_example}
\end{figure*}

{\bf Result Interpretation:}
numerical results on the test set and the external test set can be found in Table~\ref{tab:ankle_table} and the attention guidance results are visualized compared as a function of $\lambda$ and by focus region types in Fig.~\ref{fig:ankle_result}.
On the test set, the two L-norm regularization methods and the three attention mechanisms are also applied on the ankle fracture dataset, where the self attention scores $74.6\% \pm 3.1\%$, significantly improving over the baseline by a 5.6\% mean accuracy increase ($p = 0.02$), and the L2 regularization scores a marginal 3.8\% increase to $72.8\% \pm 6.7\%$ ($p = 0.29$). Whereas, the rest three methods perform worse than the baseline.
For the proposed attention guidance method on the test set, the best performance using the scribble style focus region is $80.4\% \pm 3.9\%$ at $\lambda = 0.5$, a 11.4\% improvement over the baseline. For bounding box style, it is $79.6\% \pm 3.2\%$ at $\lambda=0.5$, a 10.6\% mean accuracy improvement.
Finally, for the segmentation style, it is $80.6\% \pm 3.8\%$ at $\lambda=0.1$, a 11.6\% accuracy improvement. 

On the external test set, the effectiveness of the experimented methods show similar relative performance gain or loss to their test set counterparts. 
Finally, we note an increase above 11\% from the use of all three types of attention guidance. 
In summary, the observations on the ankle fracture dataset support the same set of conclusions from the scaphoid dataset, showing that the human attention guidance is a more effective way to improve the performance of CNN than the explored attention mechanisms. 

{\bf Visual Comparison:}
in Fig.~\ref{fig:fracture_example}, we show the visual comparison for a Weber C fracture example.
In Fig.~\ref{fig:fracture_example}-(c), the Weber A attention of the network without attention guidance is most prominent hence triggered the incorrect classification. 
Furthermore, this decision is grounded primarily to the overlaid ``L WEIGHT BEARINIG'' tag in the image and the subcutaneous soft tissue on talus and calcaneus, showing that with limited training samples, the network experienced difficulty to associate the task with the correct image pattern.
In comparison, the fracture can be correctly identified regardless of the network trained with any of the three annotation types; furthermore, the spatial self-attention normalizes attention weights between zero to one, greatly reducing effect of negative attention.
Finally, as fibula is long and thin, the focus region does not have a precise longitude guidance for the Weber classification; therefore this setting demonstrates that overestimated focus region annotation does not harm the ability of the network to explore and find supporting patterns within. We show an additional Weber B ankle fracture example in Fig.~\ref{fig:fracture_ap_example} as an error example where all the trained models, except the scribble guided models, produced a wrong prediction because the no fracture class won the competition by a large but mild activation.
Finally in Fig.~\ref{fig:fracture_9918_example}, we show a no fracture example which was misclassified as Weber type A or B likely due to the projection of the fibula over tibia.

\section{Conclusion}
In this paper, we proposed a novel CNN attention guidance method to help CNN classification networks ground decisions on more sensible visual patterns. This is achieved by modifying the state-of-the-art CNN models with self-attention and human-guided regularization. The results in the paper were reported with the ResNet-50 backbone and significant improvements in prediction accuracy were obtained on the evaluated fracture datasets, which are at typical data scales for medical image analysis problems. For completeness, we also tested with the ResNet-101 and various attentive backbones showing a consistent performance (see these supplementary experiment results in \href{https://github.com/zhibinliao89/fracture_attention_guidance}{GitHub}). 
This work signifies a step forward towards using explainable AI techniques to help humans understand and improve AI. The current method assumes a simple fracture presence and we plan to generalize it to deal with multiple, complicated, and compound fracture presences. In addition, it will be interesting to see how human annotation can be further saved in light of~\cite{wang2021towards, ji2019prsnet}.

{
	\bibliographystyle{IEEEtran}
	\bibliography{main}
}

\end{document}